\documentclass[letterpaper]{article} 
\usepackage[preprint]{aaai2027}  
\usepackage[hyphens]{url}  
\usepackage{graphicx} 
\usepackage{natbib}  
\usepackage{caption} 
\usepackage{algorithm}
\usepackage{algorithmic}

\usepackage{newfloat}
\usepackage{listings}
\DeclareCaptionStyle{ruled}{labelfont=normalfont,labelsep=colon,strut=off} 
\floatstyle{ruled}
\newfloat{listing}{tb}{lst}{}
\floatname{listing}{Listing}

\usepackage{booktabs}

\usepackage{mathptmx} 

\usepackage{cite}
\usepackage{amsmath,amssymb,amsfonts}
\usepackage{algorithmic}
\usepackage{graphicx}
\usepackage{epstopdf}
\usepackage{textcomp}
\usepackage{xcolor}
\usepackage{adjustbox}
\usepackage{color}
\usepackage{booktabs}
\usepackage{tabularx}  
\usepackage{caption}
\usepackage{threeparttable}  
\usepackage{array}  
\usepackage{multirow}  
\usepackage{color}
\usepackage{amsthm}

\usepackage{algorithm}
\usepackage{algorithmic}

\usepackage{xcolor}

\usepackage[table]{xcolor}

\usepackage{booktabs}
\usepackage{graphicx}

\usepackage{booktabs}
\usepackage{tabularx}
\usepackage{array}
\usepackage{multirow}
\usepackage{caption}
\usepackage{threeparttable}
\usepackage{xurl}
\newcolumntype{Y}{>{\raggedright\arraybackslash}X}
\newcolumntype{C}[1]{>{\centering\arraybackslash}p{#1}}
\newcolumntype{L}[1]{>{\raggedright\arraybackslash}p{#1}}

\title{Understanding and Correcting Low-Frequency Bias in EEG Foundation Model}
\author {
    Junjie Yu\textsuperscript{\rm 1,\rm 2}\equalcontrib,
    Zihan Deng\textsuperscript{\rm 1}\equalcontrib,
    Jianyu Zhang\textsuperscript{\rm 2, \rm 1, \rm 3}\equalcontrib,
    Junrong Mu\textsuperscript{\rm 4, 1, 2},
    Jiahui An\textsuperscript{\rm 5, 6, 1, 2},
    Wenxiao Ma\textsuperscript{\rm 1, 2},
    Ziling Lu\textsuperscript{\rm 1},
    Yue Wang\textsuperscript{\rm 7, 1, 2},
    Yan Zhu\textsuperscript{\rm 1},
    Kexin Lou\textsuperscript{\rm 1, 2},
    Quanying Liu\textsuperscript{\rm 1, 2}\corresponding
}
\affiliations {
    \textsuperscript{\rm 1}Department of Biomedical Engineering, Southern University of Science and Technology, Shenzhen, China\\
    \textsuperscript{\rm 2}Omni-Intelligence, Shenzhen, China\\
    \textsuperscript{\rm 3}College of Design and Engineering, National University of Singapore, Singapore\\
    \textsuperscript{\rm 4}School of Computing, National University of Singapore, Singapore\\
    \textsuperscript{\rm 5}Chinese Institute for Brain Research, Beijing, China\\
    \textsuperscript{\rm 6}Peking Union Medical College and Chinese Academy of Medical Sciences, Beijing, China\\
    \textsuperscript{\rm 7}Department of Psychology, The University of Hong Kong, Hong Kong, China\\

    liuqy@sustech.edu.cn
}

\begin{document}

\maketitle

\begin{abstract}
Increasing EEG pretraining data scale or model capacity does not consistently improve downstream performance. We identify a persistent low-frequency bias in representations learned by diverse EEG foundation models, which remains across dataset scales, model capacities, and pretraining objectives. Our analysis links this bias to the interaction between EEG's $1/f^\alpha$-like spectral structure and neural networks' tendency to preferentially learn low-frequency components. In masked autoencoders, the $\ell_2$ reconstruction objective further amplifies this imbalance: under comparable relative reconstruction errors, high-power low-frequency components contribute disproportionately to the loss. To address this issue, we introduce \textbf{FAME}, a frequency-balanced masked autoencoding framework that reconstructs time--frequency activity in predefined EEG bands from masked EEG inputs. FAME independently standardizes the reconstruction targets within each band and assigns equal weight to all band-specific losses, thereby balancing supervision across the EEG spectrum. Evaluated on 41 downstream tasks in OmniEEG-Bench, FAME learns more spectrally balanced representations and achieves state-of-the-art performance on 24 of them. These results underscore the importance of balanced spectral supervision for learning transferable EEG representations.
\end{abstract}

\section{Introduction}

Motivated by the success of large-scale pretraining in language and vision, EEG pretraining has recently emerged as an active research direction \citep{yang2023biot, gui2024eegmamba, jiang2024large, cui2024neuro}. However, existing EEG foundation models often yield only modest or inconsistent improvements over supervised learning, and increasing the scale of pretraining data or model size does not always lead to better performance \citep{yang2026eeg, liu2026eeg}. This raises a fundamental question: \emph{do larger-scale EEG datasets offer limited additional information relevant to downstream tasks, or do current pretraining paradigms fail to translate the information available at increasing data scales into more transferable representations?}

\begin{figure}[!t] 
  \centering
  \includegraphics[width=0.9\linewidth]{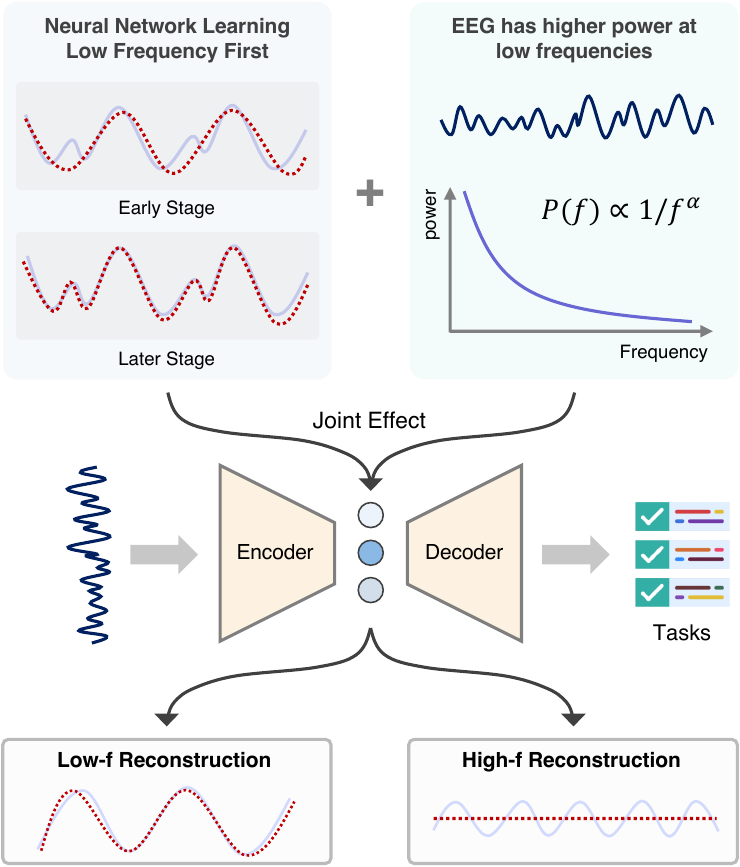}
    \caption{\textbf{Frequency bias in EEG representations.}
    Neural networks tend to fit low-frequency components first, while the
    $1/f^{\alpha}$-like EEG spectrum assigns substantially greater power to
    low-frequency activity. Their joint effect encourages the encoder to
    preferentially preserve low-frequency information.}
  \label{fig:motivation}
\end{figure}

In this paper, we focus on the latter possibility by examining an overlooked interaction between the spectral structure of EEG and the spectral bias of neural networks. EEG power typically follows a $1/f^\alpha$ distribution, with substantially greater power at lower than at higher frequencies \citep{pritchard1992brain}. Meanwhile, neural networks tend to learn low-frequency components before gradually capturing higher-frequency ones during optimization \citep{rahaman2019spectral, fridovich2022spectral}. Together, these properties may cause EEG models to preferentially encode low-frequency activity, which is both more prominent in the signal and easier to learn (Figure \ref{fig:motivation}). Through frequency-resolved analyses of existing EEG foundation models, we find that this low-frequency preference persists across pretraining objectives, dataset scales, and model capacities, indicating that scaling alone does not reliably yield more spectrally balanced representations.

Masked-reconstruction pretraining can introduce an additional, objective-level source of imbalance. By Parseval's theorem, pointwise mean squared error in the time domain is equivalent to the sum of absolute squared reconstruction errors across frequencies. Under the highly imbalanced EEG power spectrum, comparable relative reconstruction errors therefore produce much larger absolute losses at high-power, low-frequency components than at lower-power, higher-frequency components. A model can consequently reduce its reconstruction loss primarily by recovering slow, high-energy activity, while receiving substantially weaker supervision for higher-frequency dynamics. Thus, accurate signal reconstruction does not necessarily require a spectrally balanced representation, motivating an objective that equalizes learning signals across frequency bands.

To correct this objective-level imbalance, we propose \textbf{FAME}, a \textbf{F}requency-balanced \textbf{A}utoencoding framework for \textbf{M}asked \textbf{E}EG. Given masked EEG as input, FAME reconstructs time--frequency activity under a configurable partition of the EEG spectrum. Within each subband, the reconstruction targets are independently standardized, and the resulting subband-wise losses are weighted equally. This design decouples the strength of the learning signal from the intrinsic power of each subband, mitigating the dominance of high-power, low-frequency activity and encouraging the encoder to preserve information more evenly across the spectrum. In contrast, direct time--frequency reconstruction without subband-wise normalization retains the original power imbalance and therefore remains dominated by high-energy spectral components.

We evaluate FAME on \textbf{41} downstream tasks in OmniEEG-Bench \citep{lu2026omnieeg}. FAME achieves state-of-the-art performance on \textbf{24 of the 41 tasks}, demonstrating that frequency-balanced supervision produces more transferable EEG representations and improves the effectiveness of EEG pretraining.

Our contributions are threefold:
\begin{itemize}
    \item We identify a persistent low-frequency bias in EEG foundation models and show that it remains across pretraining scales, model capacities, and learning objectives.

    \item We explain this bias through the interaction between EEG's $1/f^\alpha$ spectral structure and neural networks' low-frequency learning preference.

    \item We propose \textbf{FAME}, a frequency-balanced masked autoencoding framework that promotes spectrally balanced and transferable EEG representations, achieving state-of-the-art performance on 24 of 41 tasks.
\end{itemize}

\section{Related Work}

\paragraph{Self-Supervised EEG Pretraining.}
Self-supervised EEG learning has adopted diverse objectives, including
contrastive learning, latent prediction, and masked reconstruction
\citep{banville2021uncovering, kostas2021bendr, chien2022maeeg,
wang2024eegpt, jiang2024large}. Prior work has primarily examined the
design and scaling of pretraining frameworks, including masking,
tokenization, channel alignment, and model architecture, while paying
less attention to the information retained by the resulting
representations. In particular, whether different pretraining paradigms
preserve information broadly across the EEG spectrum remains unclear.
We address this gap by systematically characterizing the
frequency-resolved information content of representative EEG foundation
models.

\paragraph{Frequency Bias and EEG Spectral Structure.}
Neural networks tend to learn low-frequency components more readily than
high-frequency ones, a phenomenon known as frequency or spectral bias
\citep{rahaman2019spectral, xu2019frequency, fridovich2022spectral}.
This tendency is especially relevant to EEG, whose power spectral density
generally decays with frequency according to an approximate
$1/f^\alpha$ relationship
\citep{pritchard1992brain, donoghue2020parameterizing,
wen2016separating}. Because low-frequency components are both
higher-powered and easier to learn, they may dominate learned
representations. However, lower-power high-frequency activity can encode
important information for motor imagery, sleep staging, and seizure
analysis
\citep{pfurtscheller1999event, silber2007visual,
bragin1999hippocampal}. Its systematic underrepresentation may therefore
remove task-relevant spectral cues and limit downstream transfer.

\paragraph{Frequency-Aware EEG Representation Learning.}
Prior work has incorporated spectral structure through Fourier-domain
tokenization \citep{yang2023biot}, masked spectrogram prediction
\citep{wang2023brainbert}, and time--frequency contrastive learning
\citep{zhang2022self}. However, explicitly modeling frequency does not
guarantee balanced supervision: in reconstruction-based pretraining,
unnormalized targets preserve large power differences across frequency
components, causing high-power components to dominate the loss even in
the time--frequency domain. FAME instead standardizes reconstruction
targets independently within each frequency subband and equally weights
the resulting subband losses. This decouples supervision from intrinsic
subband power and provides more balanced learning signals across the EEG
spectrum.


\section{Method}

We first explain why traditional mask-reconstruction provides power-weighted
spectral supervision and therefore tends to emphasize low-frequency EEG
activity. We then characterize the frequency preference of pretrained
representations from embedding similarity and information recoverability.
Based on these observations, we introduce \textbf{FAME}, a
\textbf{F}requency-balanced \textbf{A}utoencoding framework for
\textbf{M}asked \textbf{E}EG. Finally, we quantify encoder-level spectral
bias and examine its association with downstream generalization.

\subsection{Spectral Bias of Mask-Reconstruction}

By Parseval's theorem, a mask-reconstruction loss can be expressed
in the frequency domain as
\begin{equation}
    \|\mathbf{x}-\widehat{\mathbf{x}}\|_2^2
    \propto
    \sum_f
    \left|
        X_f-\widehat{X}_f
    \right|^2
    =
    \sum_f
    |X_f|^2 |r_f|^2,
    \label{eq:waveform_spectral_weighting}
\end{equation}
where $X_f$ and $\widehat{X}_f$ are the Fourier coefficients of the
target and reconstruction, and
$r_f=(X_f-\widehat{X}_f)/(|X_f|+\epsilon)$ denotes the relative spectral
error. Thus, mask-reconstruction implicitly weights relative errors
by the signal power $|X_f|^2$. Because EEG power is typically concentrated
at low frequencies, this objective provides substantially stronger
low-frequency supervision.

\subsection{Frequency Preference of Pretrained Representations}
\label{sec:frequency_preference}

We characterize frequency preference from two complementary perspectives:
the similarity between broadband and frequency-specific embeddings, and
the recoverability of frequency-specific activity from broadband
representations.

\paragraph{Embedding similarity.}

For each EEG segment $\mathbf{x}_i$, we extract broadband and
frequency-specific embeddings as
\begin{equation}
    \mathbf{z}_i
    =
    g\!\left(E_{\theta}(\mathbf{x}_i)\right),
    \qquad
    \mathbf{z}_i^{(f)}
    =
    g\!\left(E_{\theta}(\mathcal{F}_f(\mathbf{x}_i))\right),
    \label{eq:frequency_embeddings}
\end{equation}
where $E_{\theta}$ is a frozen pretrained encoder, $g$ is a fixed
token-aggregation operation, and $\mathcal{F}_f$ extracts a narrow band
centered at frequency $f$.

For each encoder, a shared PCA projection is fitted to the union of its
broadband and frequency-specific embeddings, retaining 100 components.
We then compute the Wasserstein Distance
\begin{equation}
    W_1(P_{\mathrm{full}},P_f)
    =
    \inf_{\gamma\in\Pi(P_{\mathrm{full}},P_f)}
    \mathbb{E}_{(\mathbf{u},\mathbf{v})\sim\gamma}
    \left[
        \|\mathbf{u}-\mathbf{v}\|_2
    \right],
    \label{eq:embedding_wasserstein}
\end{equation}
where $P_{\mathrm{full}}$ and $P_f$ denote the projected broadband and
frequency-specific embedding distributions. A smaller Wasserstein distance indicates
that activity around $f$ induces representations more similar to those
of the broadband signal.

\paragraph{Frequency-specific recoverability.}

For each frequency $f$, we train a separate three-layer MLP $h_f$ to
decode the corresponding activity from the broadband embedding:
\begin{equation}
    \widehat{\mathbf{y}}_i^{(f)}
    =
    h_f(\mathbf{z}_i),
    \qquad
    \mathbf{y}_i^{(f)}
    =
    \mathcal{F}_f(\mathbf{x}_i).
    \label{eq:band_decoding}
\end{equation}
After independently standardizing the targets at each frequency, the
decoding loss is
\begin{equation}
    \ell_f
    =
    \frac{1}{N}
    \sum_{i=1}^{N}
    \left\|
        h_f(\mathbf{z}_i)
        -
        \widetilde{\mathbf{y}}_i^{(f)}
    \right\|_2^2.
    \label{eq:band_decoding_loss}
\end{equation}
A smaller $\ell_f$ indicates that more information at frequency $f$ is
preserved in the broadband representation.

\paragraph{Encoder-level frequency bias.}
To quantify the spectral profile of information recoverability across
encoders, we define three complementary metrics based on the recoverability loss:
low-frequency bias, frequency imbalance, and signed frequency slope.
Let
\begin{equation}
    q_f=\log \ell_f,
    \qquad
    \overline{q}_{\mathcal{S}}
    =
    \frac{1}{|\mathcal{S}|}
    \sum_{f\in\mathcal{S}} q_f,
\end{equation}
where $\ell_f$ denotes the recoverability loss at frequency $f$.

We first define
$\mathcal{F}_{\mathrm{low}}=\{f:1\leq f<15\}$ and
$\mathcal{F}_{\mathrm{high}}=\{f:30\leq f<45\}$.
The \textbf{low-frequency bias} is
\begin{equation}
    B_{\mathrm{low}}
    =
    100
    \left[
        \exp\!\left(
            \overline{q}_{\mathcal{F}_{\mathrm{high}}}
            -
            \overline{q}_{\mathcal{F}_{\mathrm{low}}}
        \right)
        -1
    \right].
    \label{eq:low_frequency_bias}
\end{equation}
This quantity measures the percentage increase in the geometric-mean
recoverability loss from the low-frequency range to the high-frequency
range. A positive value indicates that high-frequency activity is less
recoverable than low-frequency activity.

To quantify the overall unevenness of recoverability across the analyzed
frequency set $\mathcal{F}$, we define the \textbf{frequency imbalance}
as
\begin{equation}
    I_{\mathrm{freq}}
    =
    100
    \left[
        \exp\!\left(
            \sqrt{
                \frac{1}{|\mathcal{F}|}
                \sum_{f\in\mathcal{F}}
                \left(
                    q_f-\overline{q}_{\mathcal{F}}
                \right)^2
            }
        \right)
        -1
    \right].
    \label{eq:frequency_imbalance}
\end{equation}
This direction-agnostic metric captures arbitrary variation in
frequency-wise recoverability, including non-monotonic spectral
profiles.

Finally, to characterize the direction and magnitude of the global
frequency trend, we fit
\begin{equation}
    q_f=\beta_0+\beta_1 f+\varepsilon_f
\end{equation}
and define the \textbf{signed frequency slope} as the percentage change
in fitted loss per 10~Hz:
\begin{equation}
    S_{10}
    =
    100
    \left[
        \exp(10\beta_1)-1
    \right].
    \label{eq:frequency_slope}
\end{equation}
Positive $S_{10}$ indicates decreasing recoverability with increasing
frequency, whereas negative $S_{10}$ indicates the opposite trend.

Thus, $B_{\mathrm{low}}$ measures the contrast between predefined low-
and high-frequency ranges, $I_{\mathrm{freq}}$ quantifies global
spectral unevenness regardless of direction, and $S_{10}$ captures the
signed linear trend across frequencies. For each downstream task, we
correlate these encoder-level metrics with linear-probing performance to
determine whether stronger frequency bias is associated with poorer task
performance.

\subsection{Frequency-Balanced Autoencoding for Masked EEG}

\begin{figure}[!t] 
  \centering
  \includegraphics[width=0.98\linewidth]{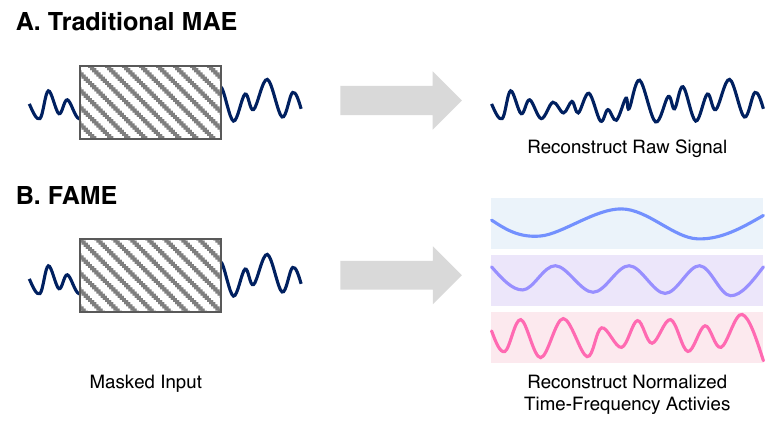}
      \caption{\textbf{Traditional reconstruction vs. frequency-balanced reconstruction.}
  \textbf{(A)} Conventional masked autoencoding reconstructs the raw EEG signal from masked inputs.
  \textbf{(B)} FAME reconstructs independently standardized time--frequency activity across predefined frequency bands with equal band-wise weighting, providing more frequency-balanced supervision.}

  \label{fig:method}
\end{figure}

FAME reconstructs band-wise standardized activity instead of the raw signal (Figure \ref{fig:method}). The EEG spectrum can be divided according to a predefined frequency partition, allowing the reconstruction granularity to be adapted to different settings. In our large-scale pretraining, we use the conventional five-band partition consisting of the $\delta$, $\theta$, $\alpha$, $\beta$, and $\gamma$ bands.

\paragraph{Masked EEG encoding.}

We divide an EEG segment $\mathbf{x}$ into $N$ non-overlapping temporal
patches and replace a random subset with mask tokens while retaining
their positions:
\begin{equation}
    \mathbf{z}
    =
    E_{\theta}
    \left(
        \{\widetilde{\mathbf{p}}_i\}_{i=1}^{N}
    \right),
    \qquad
    \widetilde{\mathbf{p}}_i
    =
    \begin{cases}
        \mathbf{p}_i, & m_i=0,\\
        \mathbf{p}_{\mathrm{mask}}, & m_i=1,
    \end{cases}
    \label{eq:masked_encoder}
\end{equation}
where $m_i\in\{0,1\}$ is the masking indicator.

\paragraph{Frequency-balanced targets.}

The original unmasked EEG is transformed into log-power
time--frequency activity:
\begin{equation}
    a_{c,f,u}
    =
    \log
    \left(
        \left|
            \mathcal{T}(\mathbf{x})_{c,f,u}
        \right|^2+\epsilon
    \right),
    \label{eq:tf_target}
\end{equation}
where $c$, $f$, and $u$ index channel, frequency bin, and temporal frame,
respectively.

Let
\begin{equation}
    \mathcal{B}=\{B_k\}_{k=1}^{K}
\end{equation}
denote a predefined partition of the analyzed EEG spectrum into $K$
frequency bands, and let $\mathcal{I}_b$ be the set of frequency bins
assigned to band $b\in\mathcal{B}$. In our large-scale pretraining,
$\mathcal{B}$ is instantiated using the conventional $\delta$, $\theta$,
$\alpha$, $\beta$, and $\gamma$ bands. We obtain one target per band by
averaging the log-power values of its constituent frequency bins:
\begin{equation}
    A_{c,b,u}
    =
    \frac{1}{|\mathcal{I}_b|}
    \sum_{f\in\mathcal{I}_b}
    a_{c,f,u}.
    \label{eq:band_aggregation}
\end{equation}
Each band is then standardized independently using corpus-level
statistics:
\begin{equation}
    \widetilde{A}_{c,b,u}
    =
    \frac{
        A_{c,b,u}-\mu_b
    }{
        \sigma_b+\epsilon
    },
    \qquad b\in\mathcal{B}.
    \label{eq:band_normalization}
\end{equation}

\paragraph{Dense frequency-balanced reconstruction.}

The decoder predicts the activity of every band at each temporal
position:
\begin{equation}
    \widehat{\mathbf{A}}
    =
    D_{\phi}(\mathbf{z}).
\end{equation}
The FAME objective is
\begin{equation}
    \mathcal{L}_{\mathrm{FAME}}
    =
    \frac{1}{|\mathcal{B}|}
    \sum_{b\in\mathcal{B}}
    \frac{1}{|\Omega_b|}
    \sum_{(c,u)\in\Omega_b}
    \left(
        \widehat{A}_{c,b,u}
        -
        \widetilde{A}_{c,b,u}
    \right)^2,
    \label{eq:fame_loss}
\end{equation}
where $\Omega_b$ contains all channel--time targets for band $b$,
including both masked and unmasked positions. Thus, masking corrupts
only the encoder input, while the decoder receives dense supervision
over the full sequence. Independent band-wise standardization mitigates
scale differences across frequency bands, while equal averaging across
bands prevents high-power bands or bands containing more frequency bins
from dominating the objective. For the experiments reported in the main
text, we use masking ratios of 50\% and 75\% for the 50M- and
1B-parameter models, respectively. Additional implementation details
are provided in the supplementary material.





\section{Experiments}

We study four questions:
(1) whether existing EEG foundation models exhibit frequency-biased representations;
(2) whether FAME alleviates such bias;
(3) whether FAME improves downstream generalization; and
(4) which downstream tasks benefit most from more frequency-balanced representations.

\subsection{Pretrained EEG Encoders Exhibit a Systematic Low-Frequency Preference}
\label{sec:spectral_bias_existing}

We examine the frequency preference of pretrained EEG representations
from two complementary perspectives: the alignment of broadband
embeddings with embeddings elicited by individual frequency bands, and
the recoverability of frequency-specific activity from broadband
representations. These analyses characterize both which spectral
components most strongly shape the representation and how much
band-specific information it retains.

\begin{figure}[!t]
  \centering
  \includegraphics[width=0.95\linewidth]{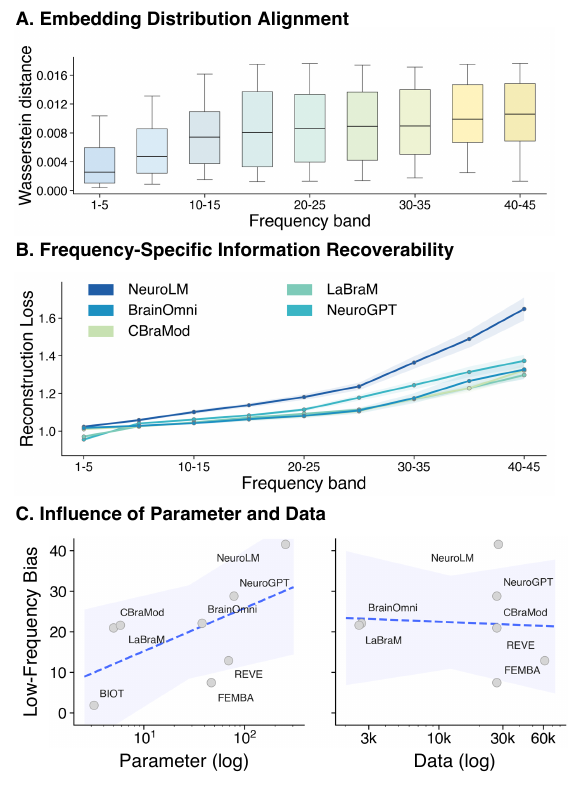}
  \caption{
  \textbf{Existing pretrained EEG encoders preferentially represent low-frequency activity.}
  \textbf{(A)} Wasserstein distances between broadband and 5-Hz band-limited embedding distributions across pretrained encoders.
  \textbf{(B)} Band-wise reconstruction from frozen NeuroGPT embeddings on TUAB.
  \textbf{(C)} Band-wise reconstruction losses under increasing pretraining data and model size, respectively.
  }
  \label{fig:frequencyBias}
\end{figure}

\paragraph{Embedding Distribution Alignment.}
We compare the embedding distribution of broadband EEG with those
obtained from consecutive 5-Hz bands spanning 0--45~Hz. For each band,
we compute its Wasserstein distance to the broadband embedding
distribution, with a smaller distance indicating stronger alignment.
As shown in Figure~\ref{fig:frequencyBias}(A), the distance generally
increases with frequency across all evaluated encoders. Broadband
representations are therefore shaped more strongly by low-frequency
activity, despite substantial differences among the models in
architecture and pretraining objective. This consistency suggests that
low-frequency preference is a shared property of existing pretrained
EEG representations rather than a consequence of a specific model or
masked-autoencoding objective.

\paragraph{Frequency-Specific Information Recoverability.}
We train separate three-layer MLP decoders to reconstruct individual
5-Hz components from broadband EEG embeddings. Across all evaluated
models, decoding loss consistently increases toward higher frequencies
(Figure~\ref{fig:frequencyBias}(B)), indicating that pretrained EEG
representations systematically retain less recoverable information about high-frequency activity.

We quantify this disparity using the frequency-bias metric derived from the band-wise decoding losses. Rather than diminishing with scale, the bias becomes more pronounced as model capacity increases (Figure~\ref{fig:frequencyBias}(C)), increasing the amount of pretraining data likewise does not consistently alleviate it. These results show that scaling alone does not yield more spectrally balanced representations and may even reinforce the preference for low-frequency information.

\subsection{Controlled Comparison between Traditional MAE and FAME}
\label{sec:mae_comparison}

We next conduct a controlled experiment to determine whether FAME mitigates the low-frequency preference induced by conventional signal reconstruction. Both models are trained for 10 epochs on single-channel EEG from TUAB using six-layer MLPs. Given a masked EEG segment of length $T$, traditional MAE reconstructs the raw signal in $\mathbb{R}^{T}$. For this controlled analysis, we instantiate FAME using nine consecutive 5-Hz bands spanning 0--45~Hz, producing a time--frequency reconstruction in $\mathbb{R}^{9T}$. The band-specific targets are independently standardized and equally weighted. The two models share the same encoder architecture and hidden dimensions, while differing in their reconstruction heads and targets.

\begin{figure}[!t]
  \centering
  \includegraphics[width=0.99\linewidth]{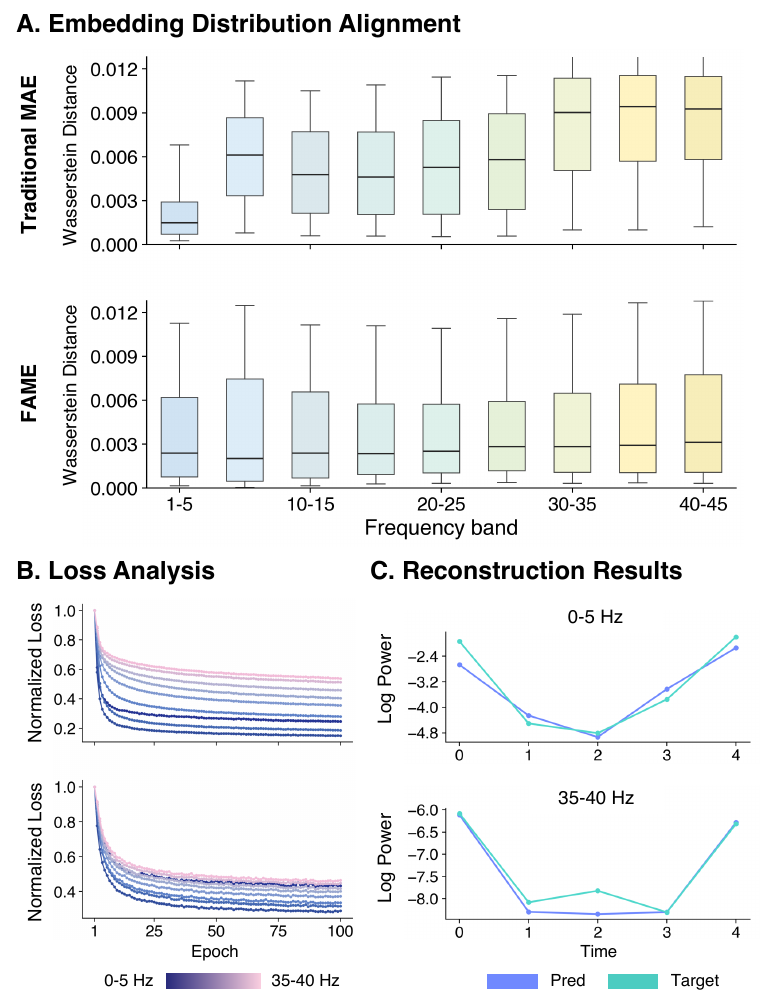}
  \caption{
  \textbf{Controlled comparison between traditional MAE and FAME.}
  \textbf{(A)} Wasserstein distances between broadband and 5-Hz band-limited embeddings extracted from the fourth MLP layer.
  \textbf{(B)} Evolution of band-wise reconstruction losses during training. Traditional MAE is dominated by low-frequency activity, whereas FAME maintains comparable losses across bands.
  \textbf{(C)} Representative targets and FAME reconstructions for different frequency bands.
  }
  \label{fig:MAEComparison}
\end{figure}

\paragraph{Representational Frequency Preference}
We extract embeddings from the fourth hidden layer of each encoder and compute the Wasserstein distance between the embedding distributions of broadband EEG and those of individual 5-Hz components. As shown in Figure~\ref{fig:MAEComparison}(A), traditional MAE yields substantially smaller distances for low-frequency components, with the distance increasing consistently toward higher frequencies. This frequency-dependent trend is markedly weaker for FAME, indicating a more spectrally balanced representation.

\definecolor{best}{RGB}{255,220,238}
\definecolor{famelarge}{RGB}{255,220,238}

\begin{table*}[!t]
\centering
\tiny
\setlength{\tabcolsep}{1.5pt}
\renewcommand{\arraystretch}{0.95}
\resizebox{\linewidth}{!}{
\begin{tabular}{l|ccccccccc|cc}
\toprule
\textbf{Dataset}
& \textbf{BIOT}
& \textbf{BrainOmni}
& \textbf{CBraMod}
& \textbf{EEGMamba}
& \textbf{FEMBA}
& \textbf{LaBraM}
& \textbf{NeuroGPT}
& \textbf{NeuroLM}
& \textbf{REVE}
& \textbf{FAME-50M}
& \textbf{FAME-1B} \\
\midrule

AD65
& 42.46 $\pm$ 8.48
& 45.48 $\pm$ 8.10
& 50.17 $\pm$ 13.86
& 37.89 $\pm$ 4.70
& 48.17 $\pm$ 11.59
& 41.05 $\pm$ 6.03
& 37.92 $\pm$ 2.35
& 33.51 $\pm$ 11.50
& 48.34 $\pm$ 13.08
& 48.67 $\pm$ 6.52
& \cellcolor{famelarge}\textbf{53.43 $\pm$ 1.87} \\

ADHD
& 54.07 $\pm$ 4.18
& 63.60 $\pm$ 8.96
& 61.95 $\pm$ 5.31
& 56.30 $\pm$ 7.84
& 59.48 $\pm$ 6.27
& 58.73 $\pm$ 8.37
& 71.06 $\pm$ 8.36
& 61.51 $\pm$ 9.84
& 62.97 $\pm$ 8.45
& 57.60 $\pm$ 5.99
& \cellcolor{famelarge}\textbf{86.39 $\pm$ 3.75} \\

Awakening
& 96.94 $\pm$ 1.89
& 96.33 $\pm$ 2.93
& 85.19 $\pm$ 6.12
& 73.91 $\pm$ 7.49
& 96.29 $\pm$ 2.06
& 95.52 $\pm$ 3.46
& 78.86 $\pm$ 3.98
& 85.48 $\pm$ 8.12
& 95.14 $\pm$ 3.01
& 92.52 $\pm$ 5.96
& \cellcolor{famelarge}\textbf{99.31 $\pm$ 0.17} \\

BCIC4-1
& 54.28 $\pm$ 7.55
& 51.65 $\pm$ 4.27
& 54.56 $\pm$ 7.84
& 48.31 $\pm$ 3.94
& 49.42 $\pm$ 1.54
& 51.52 $\pm$ 8.93
& 46.81 $\pm$ 3.69
& 49.45 $\pm$ 1.47
& 47.10 $\pm$ 11.18
& \cellcolor{best}\textbf{56.87 $\pm$ 6.26}
& 54.50 $\pm$ 2.02 \\

BCI Speech
& 21.41 $\pm$ 1.03
& 22.48 $\pm$ 1.89
& \cellcolor{best}\textbf{28.62 $\pm$ 2.23}
& 19.24 $\pm$ 1.63
& 20.85 $\pm$ 1.72
& 21.00 $\pm$ 1.12
& 19.97 $\pm$ 1.51
& 20.48 $\pm$ 1.23
& 22.44 $\pm$ 1.43
& 22.20 $\pm$ 1.67
& 23.76 $\pm$ 0.93 \\

BETA SSVEP
& 2.91 $\pm$ 0.39
& 3.53 $\pm$ 0.57
& \cellcolor{best}\textbf{11.04 $\pm$ 1.74}
& 3.01 $\pm$ 0.55
& 3.83 $\pm$ 0.44
& 3.38 $\pm$ 0.63
& 2.55 $\pm$ 0.36
& 3.32 $\pm$ 0.63
& 4.23 $\pm$ 0.61
& 7.17 $\pm$ 1.33
& 8.84 $\pm$ 2.68\\

Benchmark SSVEP
& 7.26 $\pm$ 1.92
& 8.48 $\pm$ 1.74
& 40.52 $\pm$ 9.58
& 3.09 $\pm$ 0.78
& 5.45 $\pm$ 0.85
& 3.38 $\pm$ 0.84
& 2.73 $\pm$ 0.76
& 4.08 $\pm$ 1.71
& 6.03 $\pm$ 1.06
& 25.00 $\pm$ 5.81
& \cellcolor{famelarge}\textbf{42.13 $\pm$ 12.65} \\

Broderick CP 128
& \cellcolor{best}\textbf{52.31 $\pm$ 13.05}
& 40.58 $\pm$ 12.98
& 48.04 $\pm$ 29.94
& 44.78 $\pm$ 16.42
& 44.43 $\pm$ 11.24
& 45.18 $\pm$ 13.42
& 43.29 $\pm$ 12.85
& 47.94 $\pm$ 8.81
& 46.45 $\pm$ 15.22
& 42.40 $\pm$ 5.16
& 38.34 $\pm$ 3.13 \\


CIRE
& 51.45 $\pm$ 1.18
& 50.80 $\pm$ 2.03
& 53.71 $\pm$ 5.01
& 49.93 $\pm$ 0.24
& 50.49 $\pm$ 1.25
& 49.86 $\pm$ 1.77
& 50.33 $\pm$ 2.13
& 53.15 $\pm$ 3.68
& 51.69 $\pm$ 4.80
& \cellcolor{best}\textbf{54.44 $\pm$ 0.35}
& \cellcolor{famelarge}\textbf{56.95 $\pm$ 0.99} \\

DEAP Arousal
& 51.44 $\pm$ 4.24
& 52.24 $\pm$ 3.42
& \cellcolor{best}\textbf{53.31 $\pm$ 4.60}
& 50.99 $\pm$ 1.73
& 50.94 $\pm$ 2.86
& 51.37 $\pm$ 3.23
& 50.63 $\pm$ 2.06
& 52.41 $\pm$ 3.92
& 52.39 $\pm$ 1.67
& 48.78 $\pm$ 1.00
& 52.71 $\pm$ 1.00 \\

DEAP Valence
& 52.68 $\pm$ 2.95
& 53.08 $\pm$ 2.89
& \cellcolor{best}\textbf{53.54 $\pm$ 2.95}
& 50.94 $\pm$ 1.26
& 53.32 $\pm$ 2.33
& 52.56 $\pm$ 3.05
& 52.98 $\pm$ 2.32
& 52.32 $\pm$ 2.39
& 52.52 $\pm$ 3.70
& 50.17 $\pm$ 2.73
& 52.52 $\pm$ 0.86 \\

Depression Rest
& 54.07 $\pm$ 2.85
& 50.95 $\pm$ 0.71
& 52.90 $\pm$ 5.14
& 50.00 $\pm$ 0.01
& 53.29 $\pm$ 1.92
& 51.77 $\pm$ 1.69
& 51.73 $\pm$ 2.29
& 52.30 $\pm$ 2.27
& 51.74 $\pm$ 0.65
& \cellcolor{best}\textbf{56.43 $\pm$ 5.51}
& \cellcolor{famelarge}\textbf{54.26 $\pm$ 1.42} \\

DUAL FREQ SSVEP
& 3.08 $\pm$ 0.63
& 5.16 $\pm$ 1.20
& \cellcolor{best}\textbf{8.56 $\pm$ 3.48}
& 2.78 $\pm$ 0.44
& 3.72 $\pm$ 1.22
& 2.60 $\pm$ 0.86
& 2.91 $\pm$ 0.41
& 3.01 $\pm$ 0.82
& 3.58 $\pm$ 0.64
& 3.72 $\pm$ 0.30
& 5.89 $\pm$ 0.30 \\

EAV
& 18.18 $\pm$ 6.94
& 22.07 $\pm$ 9.72
& 27.05 $\pm$ 12.41
& 20.42 $\pm$ 8.19
& 19.25 $\pm$ 6.73
& 19.62 $\pm$ 7.46
& 18.10 $\pm$ 6.60
& 18.04 $\pm$ 6.70
& 20.91 $\pm$ 9.72
& \cellcolor{best}\textbf{33.03 $\pm$ 2.75}
& \cellcolor{famelarge}\textbf{42.83 $\pm$ 2.31} \\

EEG Mortality PD
& \cellcolor{best}\textbf{65.95 $\pm$ 9.15}
& 55.75 $\pm$ 3.66
& 59.18 $\pm$ 3.37
& 50.14 $\pm$ 0.28
& 52.31 $\pm$ 4.26
& 57.95 $\pm$ 5.53
& 52.12 $\pm$ 4.31
& 57.79 $\pm$ 6.54
& 58.94 $\pm$ 7.05
& 57.86 $\pm$ 4.00
& 54.59 $\pm$ 2.07 \\

EEG SVRec
& 50.52 $\pm$ 1.04
& 49.63 $\pm$ 0.75
& 50.69 $\pm$ 3.27
& 49.94 $\pm$ 0.70
& 50.13 $\pm$ 0.31
& 51.11 $\pm$ 6.35
& 49.73 $\pm$ 0.88
& 52.32 $\pm$ 4.81
& 49.11 $\pm$ 1.46
& \cellcolor{best}\textbf{52.94 $\pm$ 2.10}
& \cellcolor{famelarge}\textbf{53.91 $\pm$ 1.03} \\

FACED
& 14.85 $\pm$ 1.13
& 21.12 $\pm$ 1.62
& \cellcolor{best}\textbf{49.98 $\pm$ 3.14}
& 14.70 $\pm$ 0.75
& 17.60 $\pm$ 1.93
& 15.44 $\pm$ 1.01
& 15.46 $\pm$ 0.66
& 18.04 $\pm$ 2.02
& 20.59 $\pm$ 1.45
& 33.00 $\pm$ 1.66
& 43.10 $\pm$ 2.23 \\

HFO
& 60.66 $\pm$ 5.16
& 56.89 $\pm$ 3.10
& 59.94 $\pm$ 4.35
& 59.33 $\pm$ 5.41
& 54.14 $\pm$ 3.73
& 56.92 $\pm$ 1.62
& 56.28 $\pm$ 3.60
& 57.04 $\pm$ 3.24
& 59.78 $\pm$ 3.96
& \cellcolor{best}\textbf{67.47 $\pm$ 8.89}
& \cellcolor{famelarge}\textbf{79.24 $\pm$ 1.19} \\

ISRUC-S1
& 51.71 $\pm$ 3.45
& 55.20 $\pm$ 3.31
& 48.47 $\pm$ 4.53
& 34.99 $\pm$ 1.57
& 53.87 $\pm$ 3.74
& 50.81 $\pm$ 2.57
& 51.59 $\pm$ 3.97
& 51.04 $\pm$ 6.80
& 57.70 $\pm$ 4.49
& \cellcolor{best}\textbf{62.96 $\pm$ 1.89}
& \cellcolor{famelarge}\textbf{61.40 $\pm$ 0.30} \\

ISRUC-S2
& 23.70 $\pm$ 1.07
& 22.96 $\pm$ 2.78
& 20.44 $\pm$ 1.91
& 20.57 $\pm$ 1.44
& 25.71 $\pm$ 2.70
& 21.20 $\pm$ 1.16
& 25.57 $\pm$ 2.57
& 21.43 $\pm$ 1.11
& 22.96 $\pm$ 1.82
& 23.82 $\pm$ 1.61
& \cellcolor{famelarge}\textbf{26.37 $\pm$ 1.13} \\

ISRUC-S3
& 25.63 $\pm$ 3.75
& 25.61 $\pm$ 2.14
& 28.72 $\pm$ 4.01
& 23.80 $\pm$ 1.54
& 24.68 $\pm$ 0.88
& 26.52 $\pm$ 2.61
& 26.07 $\pm$ 3.57
& 25.44 $\pm$ 6.20
& 24.60 $\pm$ 0.37
& 25.11 $\pm$ 1.92
& \cellcolor{famelarge}\textbf{29.10 $\pm$ 1.05} \\

LEMON Age
& 50.52 $\pm$ 7.73
& 54.06 $\pm$ 4.45
& 56.76 $\pm$ 7.99
& 33.33 $\pm$ 0.00
& 51.40 $\pm$ 6.28
& 51.32 $\pm$ 8.19
& 47.38 $\pm$ 5.92
& 51.30 $\pm$ 6.75
& 52.78 $\pm$ 2.42
& 47.42 $\pm$ 5.68
& \cellcolor{famelarge}\textbf{76.73 $\pm$ 5.98} \\

LEMON Extraversion
& 48.10 $\pm$ 0.88
& 52.43 $\pm$ 3.09
& 49.08 $\pm$ 2.48
& 49.96 $\pm$ 0.26
& 48.42 $\pm$ 1.24
& 52.45 $\pm$ 4.27
& 45.79 $\pm$ 3.05
& 50.93 $\pm$ 2.57
& 49.50 $\pm$ 2.92
& \cellcolor{best}\textbf{53.93 $\pm$ 1.93}
& 49.70 $\pm$ 0.17 \\

LEMON Gender
& 57.76 $\pm$ 1.26
& 58.32 $\pm$ 0.77
& \cellcolor{best}\textbf{71.97 $\pm$ 3.35}
& 50.56 $\pm$ 0.25
& 57.91 $\pm$ 4.13
& 59.99 $\pm$ 5.05
& 53.50 $\pm$ 5.55
& 64.38 $\pm$ 5.65
& 61.28 $\pm$ 1.18
& 61.70 $\pm$ 10.65
& 66.03 $\pm$ 6.58 \\


Monitoring ErrP
& 49.96 $\pm$ 0.96
& 52.98 $\pm$ 1.78
& 54.21 $\pm$ 4.98
& 52.27 $\pm$ 0.94
& 50.91 $\pm$ 1.80
& 51.11 $\pm$ 2.47
& 50.73 $\pm$ 2.29
& 50.85 $\pm$ 1.14
& 50.99 $\pm$ 2.34
& \cellcolor{best}\textbf{55.84 $\pm$ 4.14}
& \cellcolor{famelarge}\textbf{60.54 $\pm$ 1.20} \\

MusicEEG
& 46.88 $\pm$ 2.90
& 49.70 $\pm$ 3.74
& 53.48 $\pm$ 4.73
& 50.80 $\pm$ 2.49
& 49.10 $\pm$ 3.46
& 49.24 $\pm$ 5.11
& 49.15 $\pm$ 5.12
& 50.04 $\pm$ 2.81
& 51.49 $\pm$ 3.33
& \cellcolor{best}\textbf{54.09 $\pm$ 0.95}
& \cellcolor{famelarge}\textbf{54.53 $\pm$ 2.73} \\

PD31
& 54.62 $\pm$ 7.47
& 49.52 $\pm$ 8.67
& 46.69 $\pm$ 18.85
& 39.85 $\pm$ 21.62
& 52.82 $\pm$ 18.72
& 47.27 $\pm$ 13.96
& 46.23 $\pm$ 15.96
& 51.23 $\pm$ 25.86
& 43.65 $\pm$ 23.53
& \cellcolor{best}\textbf{55.28 $\pm$ 12.30}
& \cellcolor{famelarge}\textbf{57.04 $\pm$ 8.30} \\

PEARL Neuro
& 45.34 $\pm$ 3.68
& 48.18 $\pm$ 2.68
& 49.63 $\pm$ 11.82
& 42.98 $\pm$ 1.41
& 44.61 $\pm$ 2.86
& 47.45 $\pm$ 3.12
& 43.45 $\pm$ 2.65
& 46.68 $\pm$ 1.73
& 47.70 $\pm$ 5.82
& \cellcolor{best}\textbf{54.90 $\pm$ 4.22}
& \cellcolor{famelarge}\textbf{57.52 $\pm$ 0.42} \\

PhysioNet MI
& 25.85 $\pm$ 2.10
& 33.52 $\pm$ 4.70
& 49.77 $\pm$ 1.64
& 27.11 $\pm$ 1.38
& 29.08 $\pm$ 1.82
& 28.13 $\pm$ 3.37
& 28.92 $\pm$ 1.83
& 30.53 $\pm$ 2.50
& 34.16 $\pm$ 2.66
& 45.14 $\pm$ 4.70
& \cellcolor{famelarge}\textbf{53.86 $\pm$ 0.81} \\

RestCog
& 36.35 $\pm$ 3.07
& 35.10 $\pm$ 2.64
& \cellcolor{best}\textbf{42.19 $\pm$ 4.16}
& 25.67 $\pm$ 1.28
& 33.39 $\pm$ 2.66
& 35.26 $\pm$ 2.37
& 29.05 $\pm$ 2.27
& 32.81 $\pm$ 2.53
& 36.89 $\pm$ 2.88
& 39.68 $\pm$ 1.18
& 41.10 $\pm$ 1.34 \\

SEED
& 50.43 $\pm$ 5.28
& 51.35 $\pm$ 5.94
& 58.71 $\pm$ 3.48
& 47.13 $\pm$ 4.48
& 51.62 $\pm$ 8.36
& 48.97 $\pm$ 4.48
& 45.06 $\pm$ 4.33
& 49.84 $\pm$ 3.43
& 56.88 $\pm$ 7.25
& \cellcolor{best}\textbf{60.30 $\pm$ 1.31}
& 57.15 $\pm$ 2.51 \\

SEED-IV
& 26.52 $\pm$ 5.02
& \cellcolor{best}\textbf{37.86 $\pm$ 0.74}
& 32.16 $\pm$ 1.51
& 27.40 $\pm$ 2.25
& 31.26 $\pm$ 2.90
& 34.94 $\pm$ 1.67
& 28.64 $\pm$ 1.51
& 32.08 $\pm$ 2.76
& 34.65 $\pm$ 2.89
& 35.02 $\pm$ 4.38
& 35.04 $\pm$ 1.00 \\


SEED-V
& 24.18 $\pm$ 2.62
& 26.48 $\pm$ 3.65
& 27.39 $\pm$ 4.46
& 20.55 $\pm$ 0.89
& 25.67 $\pm$ 3.31
& 23.07 $\pm$ 2.03
& 23.17 $\pm$ 1.26
& 23.75 $\pm$ 2.29
& 26.19 $\pm$ 3.97
& 25.29 $\pm$ 2.94
& \cellcolor{famelarge}\textbf{28.03 $\pm$ 1.83} \\


SEED-VII
& 15.79 $\pm$ 1.68
& \cellcolor{best}\textbf{20.47 $\pm$ 1.16}
& 18.49 $\pm$ 1.84
& 15.98 $\pm$ 2.30
& 18.37 $\pm$ 2.45
& 17.36 $\pm$ 1.45
& 16.57 $\pm$ 1.87
& 16.68 $\pm$ 1.84
& 18.72 $\pm$ 3.10
& 16.84 $\pm$ 1.09
& 18.77 $\pm$ 1.68 \\

SHU-MI
& 50.30 $\pm$ 1.69
& 53.85 $\pm$ 1.84
& 58.50 $\pm$ 4.43
& 50.81 $\pm$ 3.11
& 52.72 $\pm$ 5.68
& 49.94 $\pm$ 1.23
& 50.60 $\pm$ 1.77
& 52.09 $\pm$ 0.84
& 54.53 $\pm$ 3.11
& 57.48 $\pm$ 1.11
& \cellcolor{famelarge}\textbf{62.97 $\pm$ 0.34} \\

Siena EEG
& 67.00 $\pm$ 28.26
& 66.32 $\pm$ 28.29
& \cellcolor{best}\textbf{75.28 $\pm$ 24.46}
& 66.66 $\pm$ 28.86
& 66.65 $\pm$ 28.85
& 66.16 $\pm$ 28.01
& 66.64 $\pm$ 28.82
& 73.30 $\pm$ 29.11
& 62.07 $\pm$ 20.22
& 61.96 $\pm$ 10.91
& 51.64 $\pm$ 1.66 \\

SSVEP
& 0.62 $\pm$ 0.00
& 1.63 $\pm$ 0.61
& 2.81 $\pm$ 0.91
& 1.12 $\pm$ 0.51
& 0.74 $\pm$ 0.40
& 0.75 $\pm$ 0.21
& 0.71 $\pm$ 0.34
& 1.23 $\pm$ 0.33
& 4.43 $\pm$ 1.90
& 3.54 $\pm$ 2.04
& \cellcolor{famelarge}\textbf{8.85 $\pm$ 2.36} \\

TDBRAIN
& 38.38 $\pm$ 3.96
& 39.76 $\pm$ 0.10
& 41.49 $\pm$ 3.20
& 31.22 $\pm$ 0.71
& 40.44 $\pm$ 3.31
& 39.97 $\pm$ 3.61
& 36.28 $\pm$ 1.84
& 38.60 $\pm$ 0.60
& 30.11 $\pm$ 0.38
& 40.54 $\pm$ 1.77
& \cellcolor{famelarge}\textbf{48.61 $\pm$ 5.72} \\

TUAB
& 77.29 $\pm$ 2.20
& 74.19 $\pm$ 3.98
& 73.07 $\pm$ 1.80
& 68.92 $\pm$ 3.08
& 72.28 $\pm$ 6.54
& 74.78 $\pm$ 5.80
& 66.84 $\pm$ 3.54
& 72.35 $\pm$ 1.03
& \cellcolor{best}\textbf{78.41 $\pm$ 0.23}
& 67.87 $\pm$ 1.67
& 75.41 $\pm$ 1.86 \\

TUEV
& 66.25 $\pm$ 9.12
& 50.58 $\pm$ 2.35
& 62.56 $\pm$ 6.50
& 31.55 $\pm$ 7.60
& 46.48 $\pm$ 7.29
& 42.88 $\pm$ 12.93
& 31.33 $\pm$ 2.25
& 27.71 $\pm$ 1.95
& 53.47 $\pm$ 0.03
& \cellcolor{best}\textbf{66.95 $\pm$ 12.63}
& \cellcolor{famelarge}\textbf{80.70 $\pm$ 4.68} \\

Workload
& 28.29 $\pm$ 7.84
& 29.90 $\pm$ 4.93
& 16.00 $\pm$ 1.52
& 38.78 $\pm$ 2.17
& 31.49 $\pm$ 10.81
& 40.74 $\pm$ 4.63
& 29.03 $\pm$ 12.14
& 36.50 $\pm$ 3.37
& 9.27 $\pm$ 4.61
& 39.17 $\pm$ 2.08
& \cellcolor{famelarge}\textbf{43.62 $\pm$ 7.67} \\

\bottomrule
\end{tabular}%
}
\caption{\textbf{Linear-probing performance across downstream EEG datasets.} 
Balanced accuracy (BACC, \%, mean $\pm$ standard deviation) is reported,
with higher values indicating better performance. Results achieving
state-of-the-art performance are highlighted. When both FAME-50M and
FAME-1B outperform the best existing baseline on a dataset, both results
are highlighted.}
\label{tab:downstream_linear_probe}
\end{table*}

\paragraph{Band-Wise Optimization Dynamics.}
We further track the reconstruction error of each frequency band
throughout pretraining. For the traditional MAE, we decompose both the
target and reconstructed singals into the same nine 5-Hz bands and
compute the error separately for each band. For FAME, the corresponding
errors are obtained directly from its band-specific outputs. To remove
differences in absolute loss scale and facilitate comparison of relative
convergence rates, each band-wise loss is normalized by its initial
value.

As shown in Figure~\ref{fig:MAEComparison}(B), the normalized loss
trajectories of the conventional MAE differ substantially across
frequency bands. Low-frequency losses decrease rapidly, whereas
higher-frequency losses become progressively more difficult to reduce.
In contrast, FAME produces more closely aligned trajectories, with
losses across frequency bands decreasing at more comparable rates.
These results indicate that conventional signal reconstruction
exhibits strongly frequency-dependent optimization dynamics, whereas
FAME promotes more balanced learning across the spectrum.

\paragraph{Visualization of Reconstruction.}
Figure~\ref{fig:MAEComparison}(C) visualizes the time series
reconstructed by FAME for different frequency bands within a masked
patch, alongside the corresponding ground-truth signals. Across both
low- and high-frequency bands, the reconstructed activity closely
matches the ground truth, accurately capturing its temporal dynamics.

Overall, this controlled comparison demonstrates that conventional
signal reconstruction induces a pronounced low-frequency preference
in both optimization dynamics and learned representations. By balancing
supervision across frequency bands, FAME substantially mitigates this
bias and promotes more spectrally balanced representations.

\subsection{Generalization Performance}
\label{sec:generalization}

Our analysis above shows that FAME mitigates the intrinsic frequency
bias of EEG pretraining and enables the encoder to capture a more
balanced spectrum of neural activity. We next investigate whether such
frequency-balanced representations translate into improved
generalization across diverse downstream tasks. To this end, we pretrain
two Transformer encoders with 50M and 1B parameters and evaluate their
transferability through linear probing on 41 downstream EEG tasks.
These tasks cover a broad range of applications, such as clinical diagnosis, cognitive state recognition, and motor imagery. Further details on the pretraining data, downstream datasets, and adaptation and evaluation protocols are provided in the supplementary material.

As shown in Table~\ref{tab:downstream_linear_probe}, the 50M FAME model already achieves state-of-the-art performance on 14 out of 41 tasks,
demonstrating strong transferability. Scaling the model from 50M to 1B parameters further improves downstream performance, with the 1B model achieving state-of-the-art results on 24 tasks, including substantial gains on ADHD, HFO and TUEV dataset.
This scaling behavior further indicates that
frequency-balanced supervision enables larger models to use their
increased capacity to encode additional transferable information.

To assess generalization on an independent benchmark, we compare FAME
with seven other models across 13 tasks from Meta's NeuralBench
\citep{banville2026neuralbench}, using full-parameter fine-tuning for all
models. FAME again achieves the best average performance across the 13
tasks, showing that its advantage extends beyond linear probing and
remains significant under full-parameter fine-tuning. Detailed results
are provided in the supplementary material.


\subsection{Relationship between Frequency Bias and Generalization}
\label{sec:bias_generalization}

Motivated by the downstream gains of frequency-balanced pretraining, we
next investigate whether the spectral characteristics of pretrained
representations are systematically related to their transfer
performance. For each model, we quantify frequency bias using the three
metrics defined in Methods: low-frequency bias, frequency imbalance,
and signed frequency slope. These metrics capture complementary aspects
of how information is distributed and preserved across frequencies. We
then correlate each metric with linear-probing performance separately
for each downstream task.

\begin{figure}[htbp]
  \centering
  \includegraphics[width=0.99\linewidth]{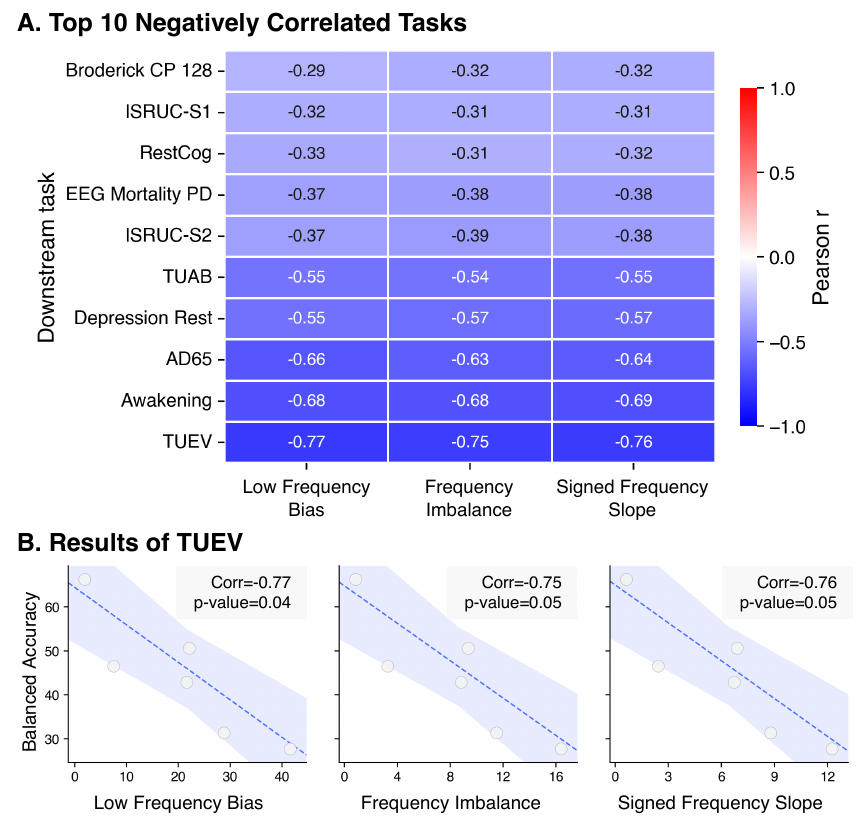}
  \caption{
  \textbf{Association between frequency bias and downstream generalization.}
  \textbf{(A)} The ten downstream tasks showing the strongest negative correlations between low-frequency bias and linear-probing performance.
  \textbf{(B)} Relationships between linear-probing performance on the TUEV and three representation metrics: low-frequency bias, frequency imbalance, and signed frequency slope. Each point denotes a model.
  }
  \label{fig:freqBias_Generalization}
\end{figure}


The relationship between frequency bias and performance varies
across tasks. Figure~\ref{fig:freqBias_Generalization}(A) highlights the
ten tasks exhibiting the strongest negative associations, while results
for all tasks are reported in the supplementary material. On these
tasks, models with less frequency-biased representations tend to achieve
better linear-probing performance, suggesting that preserving
information more evenly across the spectrum can facilitate downstream
transfer.

Figure~\ref{fig:freqBias_Generalization}(B) illustrates this pattern on
the TUEV dataset by relating linear-probing performance to each of the
three frequency-bias metrics. All three analyses show a negative
association, with the strongest correlation reaching $r=-0.77$.

Taken together, these findings indicate that frequency balance is an
important representation property for tasks that depend on information
distributed across multiple frequency ranges. Its benefit is not
universal, however, and depends on the spectral demands of the
downstream task. Tasks driven predominantly by low-frequency activity
may derive little benefit from preserving high-frequency information,
and a low-frequency preference may therefore be neutral or even
advantageous.

\section{Discussion and Conclusion}

Our results reveal a persistent low-frequency bias in current EEG
foundation models across pretraining objectives, data scales, and model
capacities, indicating that scaling alone does not guarantee more
comprehensive representations. We link this bias to EEG's
$1/f^\alpha$ spectral structure and neural networks' preference for
low-frequency patterns. In masked autoencoders, reconstruction losses
further reinforce the bias by overweighting high-power components,
allowing accurate reconstruction without adequately capturing weaker
high-frequency dynamics.

FAME mitigates this objective-level imbalance by independently
standardizing targets within frequency bands and equally weighting
band-specific losses. Rather than suppressing low-frequency information,
it prevents dominant spectral components from overwhelming supervision.
At the 1B-parameter scale, FAME achieves state-of-the-art performance on
24 of 41 OmniEEG-Bench tasks.

The benefits of frequency balancing remain task dependent. Reduced bias
is associated with better transfer to tasks requiring broader spectral
information, but may offer less benefit when slow activity predominates.
Frequency balancing should therefore be viewed as a means of learning
less spectrally constrained representations, rather than as a universally
optimal spectral prior.

\paragraph{Limitations.}
FAME requires the computation and storage of time--frequency targets,
introducing additional preprocessing and storage costs. Its predefined
frequency partition and equal band weighting provide a simple,
task-agnostic solution, but may not match the spectral characteristics
of every dataset or downstream task. 
In addition, our analysis of the
relationship between frequency bias and downstream generalization is
based on a limited number of pretrained models. The observed association
should therefore be regarded as exploratory rather than conclusive, as
the current sample size does not provide sufficient statistical power
for rigorous inference.

In summary, we show that EEG spectral structure, neural network
preferences, and reconstruction objectives jointly shape biased
representations. By balancing supervision across frequency bands, FAME
improves spectral coverage and transferability of EEG foundation models,
highlighting that effective scaling requires not only larger models and
more data, but also better-aligned pretraining objectives.

\section*{Author contributions}

J.Y. conceived the study, developed the overall framework for understanding the limitations of current EEG foundation models from the perspective of low-frequency representational bias, and proposed the FAME framework. J.Y. also pretrained the 50M-parameter model and drafted the manuscript. Z.D. performed the fine-tuning experiments for the 1B-parameter model. J.Z. processed all datasets. J.Z. and J.M. jointly fine-tuned the 50M-parameter model and analyzed its performance across all downstream tasks. J.A. conducted the analyses of low-frequency bias in the learned representations. W.M. pretrained the 1B-parameter model. Z.L. fine-tuned the comparison models across the downstream tasks. Y.W., Y.Z., K.L., and Q.L. processed the data and contributed to improving the manuscript. All authors critically revised the manuscript and contributed to its important intellectual content.

\bibliography{aaai2027}


\newpage

\appendix






\section{Details of Pretraining Datasets}
\label{app:pretraining_datasets}


This section provides detailed information about the EEG datasets used for
pretraining. We first summarize the standardized dataset-level metadata,
including the experimental paradigm, number of subjects, number of recording
sessions, number of classes or experimental conditions, channel configuration,
sampling rate, and recording amount. We then provide a concise description of
the experimental paradigm associated with each dataset.

The information reported in this section was extracted from the documentation
distributed with the corresponding datasets. A value is reported as
\textit{n/a} when the relevant information was not explicitly specified in the
available documentation. Runs, blocks, phases, rounds, and trials were not
interpreted as recording sessions unless they were explicitly described as
sessions. Similarly, the number of classes was reported only when the dataset
documentation defined an unambiguous set of labels, states, conditions, or
stimulus concepts. Experimental phases and participant groups were not
automatically treated as classification classes.

\subsection{Dataset Summary}
\label{app:dataset_summary}

Tables~\ref{tab:pretraining_datasets} summarize the principal characteristics of
the pretraining datasets. Because the datasets differ substantially in their
experimental designs, the ``Recording amount'' column reports the most
informative quantity available for each dataset, such as the number of trials,
recording duration, blocks, or sessions. Therefore, values in this column
should not necessarily be interpreted as directly comparable measures of
dataset size.

\begin{table*}[!t]
    \centering
    \begin{threeparttable}
    \setlength{\tabcolsep}{2.8pt}
    \renewcommand{\arraystretch}{1.08}
    \scriptsize

    \begin{tabularx}{\textwidth}{
        L{1.15cm}
        Y
        C{0.90cm}
        C{0.90cm}
        C{1.10cm}
        C{1.15cm}
        C{1.15cm}
        L{2.75cm}
    }
        \toprule
        \textbf{Dataset}
        & \textbf{Task / paradigm}
        & \textbf{Subjects}
        & \textbf{Sessions}
        & \textbf{Classes}
        & \textbf{Channels}
        & \textbf{Sampling rate}
        & \textbf{Recording amount} \\
        \midrule

        ds3690
        & Auditory cued reaction-time tasks, including passive listening,
          simple reaction, and go/no-go conditions
        & 75
        & n/a
        & n/a
        & 64-channel system\tnote{a}
        & 500 Hz
        & A 4-min passive task and two 8-min runs for each active task;
          240 active-task trials per subject \\

        ds3825
        & Visual object and concept perception using rapid serial visual
          presentation
        & 50
        & n/a
        & 1,854 concepts
        & 64
        & 1,000 Hz
        & Approximately 1 h per subject; more than 25,000 visual trials
          per subject \\

        ds3846
        & Haptic prediction-error and oddball task with visual,
          vibrotactile, and electrical muscle stimulation
        & 19\tnote{b}
        & n/a
        & 2 trial types\tnote{c}
        & 63 EEG + 1 EOG
        & 1,000 Hz\tnote{d}
        & 3,300 task trials from 11 subjects in the reported study \\

        ds3885
        & Static-image viewing with passive RSVP and aliveness
          categorization
        & 24
        & n/a
        & 2
        & 128
        & 1,000 Hz
        & Eight blocks per subject; four passive-viewing and four
          categorization blocks \\

        ds4043
        & Visual attention and temporal-expectation task using rapid
          overlaid grating sequences
        & 20
        & n/a
        & n/a
        & 64
        & 1,000 Hz\tnote{e}
        & 64 sequences per subject and 6,656 analyzed trials per subject \\

        ds4148
        & Resting-state and cognitive-state EEG, including subtraction,
          music listening, and memory
        & 60
        & 3
        & 5
        & 64
        & 500 Hz
        & Three recording sessions per subject \\

        ds4152
        & Motor learning and sensorimotor timing during drumming-pattern
          practice
        & 21
        & n/a
        & 12 condition codes
        & 31
        & 1,000 Hz
        & 72 trials per subject \\

        ds4264
        & Visuomotor ship-steering task under controller and environmental
          noise
        & 21
        & n/a
        & 4 noise conditions
        & n/a
        & n/a
        & n/a \\

        ds4284
        & Value-based decision-making based on descriptions of
          crowdfunding projects
        & n/a
        & n/a
        & 2
        & n/a
        & n/a
        & Funding or non-funding decision for each project \\

        ds4315
        & Reinforcement-learning task following sad or neutral mood
          manipulation
        & 50
        & n/a
        & n/a\tnote{f}
        & 66\tnote{g}
        & 500 Hz
        & Approximately 25 min per subject \\

        ds4356
        & Auditory brainstem responses to click, music, and speech stimuli
        & 24\tnote{h}
        & n/a
        & 12 stimulus types\tnote{i}
        & 34
        & 10,000 Hz
        & Ten 1-min click trials and 40 trials for each of the 12
          music/speech stimulus types \\

        ds4357
        & Visual feature coding using RSVP presentation of Gabor-like
          stimuli and fixation-color-change detection
        & 16
        & n/a
        & n/a
        & 128\tnote{j}
        & 1,000 Hz
        & n/a \\

        \bottomrule
    \end{tabularx}

    \begin{tablenotes}[flushleft]
        \footnotesize

        \item[a]
        FP1, FPz, and FP2 were excluded. The dataset additionally contains
        vertical and horizontal EOG, bipolar ECG, and pupil channels.

        \item[b]
        The dataset summary lists 19 subjects; completeness may vary across
        subjects.

        \item[c]
        The two principal trial types are oddball and non-oddball. Multiple
        stimulation conditions are also present.

        \item[d]
        The original sampling rate is 1,000 Hz; the preprocessed data are
        provided at 250 Hz.

        \item[e]
        The recordings were originally sampled at 1,000 Hz and were
        downsampled to 250 Hz in the reported preprocessing pipeline.

        \item[f]
        The documentation reports two mood-manipulation groups, but these
        groups were not treated as task classes.

        \item[g]
        The recordings contain 60 EEG channels among 66 recorded channels.

        \item[h]
        Twenty-four subjects were originally recruited, and 22 remained
        after the reported exclusions.

        \item[i]
        The 12 stimulus types consist of six music and six speech stimuli.
        A separate click-stimulation phase is also included.

        \item[j]
        The dataset documentation reports 128 channels, whereas some
        distributed dataset files appear to contain 63 channels. This
        discrepancy should be considered when processing the data.

    \end{tablenotes}
    \end{threeparttable}
    \caption{
        \textbf{Summary of the EEG datasets used for pretraining}.
        The number of channels refers to EEG channels unless otherwise
        indicated. ``n/a'' denotes information that was not explicitly
        available in the dataset documentation.
    }
    \label{tab:pretraining_datasets}
\end{table*}

\subsection{Dataset Descriptions}
\label{app:dataset_descriptions}

Rather than repeating the numerical information in
Tables~\ref{tab:pretraining_datasets}, this subsection describes the experimental content of the datasets. The datasets are grouped by their dominant
experimental modality or cognitive paradigm. These groups are used only to
organize the presentation and do not constitute labels used during
pretraining.

\subsubsection{Visual Perception and Attention}

\paragraph{ds3825: THINGS-EEG \citep{grootswagers2022human}.}
This dataset investigates neural responses to a large and diverse set of
visual object concepts. Participants viewed object images embedded in rapid
serial visual presentation streams while monitoring occasional changes in
fixation color. The dataset covers 1,854 object concepts and contains more
than 25,000 visual trials per participant. It therefore provides both
large-scale trial-level variability and broad semantic coverage, making it
particularly useful for learning transferable EEG representations of visual
processing.

\paragraph{ds3885: Object aliveness \cite{ds003885:1.0.8}.}
Participants viewed static images in alternating passive-viewing and
categorization blocks. During the categorization task, they judged whether
the presented object was alive or not alive, whereas the passive condition
required viewing the same general type of visual material without explicit
semantic categorization. The dataset consequently combines perceptual and
task-dependent responses to object images.

\paragraph{ds4043: Visual attention and temporal expectation \citep{ds004043:1.1.0}.}
This dataset was designed to dissociate the temporal effects of attention
from decision-making, memory, and expectation. Participants detected target
gratings presented within rapid sequences of overlaid visual gratings. The
large number of experimental sequences and analyzable non-target trials
provides dense temporal observations of visually evoked and attention-related
EEG activity.

\paragraph{ds4357: Visual feature coding \citep{ds004357:1.0.1}.}
This dataset examines the temporal dynamics of low-level visual feature
coding and perceptual integration. Participants viewed rapidly presented
Gabor-like visual stimuli while performing a fixation-color-change detection
task. The paradigm provides controlled variation in visual features and is
therefore complementary to datasets based on natural object images.

\subsubsection{Auditory Processing}

\paragraph{ds3690: Auditory cued reaction time \citep{ds003690:1.0.0}.}
This dataset contains EEG recordings from younger and older adults during
passive auditory listening, cued simple reaction-time, and cued go/no-go
tasks. The combination of passive and active conditions captures auditory
sensory processing, response preparation, behavioral inhibition, and
age-related variability within a common experimental framework.

\paragraph{ds4356: Music and speech auditory responses \citep{ds004356:2.2.1}.}
This dataset measures auditory brainstem and cortical responses to click,
music, and speech stimuli. The stimulus set contains six music and six speech
types, together with a separate click-stimulation phase. Its high sampling
rate preserves fast auditory responses that may not be represented in
conventional lower-sampling-rate EEG datasets.

\subsubsection{Motor, Haptic, and Visuomotor Interaction}

\paragraph{ds3846: Haptic prediction error \citep{ds003846:2.0.2}.}
This dataset uses an oddball-style paradigm to study prediction errors during
haptic and multisensory stimulation. The experimental conditions include
visual, vibrotactile, and electrical muscle stimulation, and the principal
trial distinction is between oddball and non-oddball events. The dataset
introduces somatosensory and multisensory neural dynamics into the
pretraining corpus.

\paragraph{ds4152: Drum Trainer \citep{ds004152:1.1.2}.}
Participants practiced drumming patterns in a motor-learning and
sensorimotor-timing paradigm. The task includes multiple condition codes and
repeated trials, allowing the recordings to capture motor planning, rhythmic
timing, action execution, and practice-related changes in neural activity.

\paragraph{ds4264: Steer the Ship \citep{ds004264:1.1.0}.}
This dataset was collected during a visuomotor control task in which
participants steered a virtual ship under different sources and levels of
noise. Noise could arise from the controller, the environment, or their
combination. The paradigm emphasizes continuous sensorimotor integration and
adaptive control rather than discrete stimulus classification.

\subsubsection{Resting State, Cognitive State, and Decision-Making}

\paragraph{ds4148: Test--retest resting and cognitive states \citep{ds004148:1.0.1}.}
This dataset contains repeated recordings of resting and task-related brain
states. The five states comprise eyes-closed rest, eyes-open rest, mental
subtraction, music listening, and memory-related activity. Each participant
completed three sessions, enabling the dataset to represent both
within-subject state changes and cross-session variability.

\paragraph{ds4284: EEG neuroforecasting \citep{ds004284:1.0.0}.}
Participants viewed images and textual descriptions of independent-film
projects from a crowdfunding platform and decided whether they would fund
each project. The resulting binary funding decisions provide a
value-based decision-making paradigm involving visual processing, language
comprehension, preference formation, and choice.

\paragraph{ds4315: Mood manipulation and reinforcement learning \citep{ds004315:1.0.0}.}
This dataset examines reinforcement-learning behavior following either sad
or neutral mood manipulation. The mood conditions define participant groups
rather than task classes. The recordings provide neural activity associated
with affective context, feedback processing, learning, and sequential
decision-making.

\subsection{Dataset Harmonization}
\label{app:dataset_harmonization}

The pretraining corpus combines datasets collected with heterogeneous
experimental designs, channel configurations, sampling rates, and recording
durations. These differences are intrinsic to large-scale multi-dataset EEG
pretraining. Before constructing model inputs, each dataset was processed
using the common preprocessing and tokenization pipeline described in
Section Method in the main text. Dataset-specific class labels were not required by
the self-supervised pretraining objective unless otherwise stated.

The tabulated channel counts and sampling rates describe the recordings as
distributed or as explicitly documented by the dataset authors. When both
original and preprocessed sampling rates were available, both values are
reported in the table notes. Likewise, non-EEG channels, such as EOG, ECG,
EMG, or pupil channels, are identified separately whenever this information
was available. Any inconsistency between the documentation and the
distributed files is explicitly noted rather than silently resolved.

Overall, the pretraining collection spans visual, auditory, somatosensory,
motor, cognitive-state, and decision-making paradigms. This diversity exposes
the model to neural activity arising from substantially different stimulus
modalities, behavioral requirements, recording environments, and subject
populations, thereby supporting the learning of representations that can be
transferred across heterogeneous downstream EEG tasks.

\section{Overview of Downstream Datasets}
\label{app:downstream_datasets}


We evaluate the pretrained models on 44 downstream EEG classification datasets
covering clinical diagnosis, affective and cognitive state recognition, motor
imagery, speech and auditory processing, sleep staging, event detection, and
SSVEP decoding. As summarized in Table~\ref{tab:downstream_datasets}, the
benchmark spans heterogeneous recording montages (6--128 channels), input
segment lengths (1--30 s), subject cohorts (2--1,259 subjects), and binary to
fine-grained multiclass tasks (2--160 classes). This diversity enables the
evaluation of representation transfer across distinct neural processes,
recording configurations, and data regimes.

\begin{table*}[!t]
\centering
\scriptsize
\setlength{\tabcolsep}{6pt}
\renewcommand{\arraystretch}{0.82}
\begin{tabular}{@{}lrrrr>{\raggedright\arraybackslash}p{10.0cm}@{}}
\toprule
Dataset & Subjects & Channels & Classes & Length & Task \\
\midrule
AD65 & 88 & 19 & 3 & 2 s & Neurodegenerative disease classification \\
ADHD & 121 & 19 & 2 & 2 s & Healthy vs. ADHD classification \\
Awakening & 21 & 62 & 2 & 1 s & Awake vs. sedation-state classification \\
BCI\_Speech & 25 & 64 & 5 & 3 s & Speech imagery classification \\
BCIC4\_1 & 7 & 59 & 2 & 1 s & Motor imagery classification \\
BenchmarkSSVEP & 21 & 60 & 40 & 2 s & SSVEP frequency classification \\
Broderick\_CP\_128 & 33 & 128 & 2 & 1 s & Left--right auditory attention classification \\
Broderick\_Rev\_128 & 25 & 128 & 2 & 1 s & Natural vs. time-reversed speech classification \\
CIRE & 38 & 128 & 2 & 2 s & Emotion/prosody classification \\
DEAP\_arousal & 14 & 32 & 2 & 10 s & Emotion arousal classification \\
DEAP\_valence & 32 & 32 & 2 & 10 s & Valence/preference classification \\
Depression\_rest & 121 & 66 & 2 & 1 s & Depression severity grouping based on BDI \\
DUAL\_FREQ\_SSVEP & 14 & 64 & 40 & 1 s & Dual-frequency SSVEP classification \\
EAV & 41 & 30 & 10 & 10 s & Conversational emotion classification \\
EEG\_Mortality\_PD & 76 & 63 & 2 & 1 s & Mortality vs. survival classification \\
EEG\_SVRec & 27 & 67 & 2 & 2 s & Valence/preference classification \\
FACED\_new & 123 & 30 & 9 & 10 s & Fine-grained emotion classification \\
HFO & 29 & 18 & 2 & 1 s & High-frequency-oscillation classification \\
ISRUC\_S1 & 100 & 6 & 5 & 30 s & Sleep-stage classification \\
ISRUC\_S2 & 8 & 6 & 5 & 30 s & Sleep-stage classification \\
ISRUC\_S3 & 10 & 6 & 5 & 30 s & Sleep-stage classification \\
LEMON\_age & 201 & 59 & 4 & 1 s & Age-group classification \\
LEMON\_extraversion & 6 & 59 & 2 & 1 s & Extraversion classification \\
LEMON\_gender & 201 & 59 & 2 & 1 s & Gender classification \\
MDD & 64 & 22 & 2 & 1 s & Healthy vs. major depressive disorder classification \\
MonitoringErrP & 6 & 64 & 2 & 1 s & Error-related-potential classification \\
MusicEEG & 31 & 19 & 2 & 1 s & Music-evoked emotion classification \\
PD31 & 31 & 40 & 2 & 1 s & Healthy vs. Parkinson's disease classification \\
PEARL\_Neuro & 79 & 127 & 2 & 1 s & Context/task discrimination classification \\
Physionet\_MI & 106 & 64 & 4 & 1 s & Motor imagery classification \\
RestCog & 59 & 61 & 5 & 1 s & Task-type classification \\
SEED & 15 & 62 & 3 & 10 s & Video-elicited emotion classification \\
SEED\_FRA & 8 & 60 & 3 & 10 s & Emotion classification with French movie stimuli \\
SEED\_V & 15 & 62 & 5 & 10 s & Audio-visual emotion classification \\
SEED\_VIG & 23 & 16 & 3 & 8 s & Vigilance-state classification \\
SEED\_VII & 20 & 62 & 7 & 10 s & Audio-visual emotion classification \\
SEEDIV & 15 & 62 & 4 & 10 s & Emotion classification with eye tracking \\
SHU\_MI & 18 & 30 & 2 & 1 s & Motor imagery classification \\
Siena\_EEG & 2 & 29 & 2 & 10 s & EEG abnormality detection \\
SSVEP & 13 & 9 & 160 & 3 s & SSVEP frequency classification \\
TDBRAIN & 1,259 & 26 & 4 & 2 s & Psychiatric phenotype classification \\
TUAB & 325 & 21 & 2 & 10 s & Clinical normal vs. abnormal EEG classification \\
TUEV & 449 & 26 & 6 & 5 s & EEG event classification \\
Workload & 6 & 61 & 3 & 2 s & Cognitive workload classification \\
\bottomrule
\end{tabular}
\caption{\textbf{Summary of downstream datasets and classification tasks}. The segment length denotes the input duration used for each sample.}
\label{tab:downstream_datasets}
\end{table*}

\section{Overview of the Compared Pretrained Models}
\label{app:compared_models}


Table~\ref{tab:pretrained_model_comparison} summarizes the pretrained models
considered in this study. The models differ in signal modality, pretraining
scale, tokenization strategy, self-supervised objective, backbone architecture,
and channel-handling mechanism. We include both EEG-specific models and models
developed for broader biosignal or intracranial neural-signal settings, as
these differences may affect transferability and representational properties.

For clarity, ``NR'' denotes information that was not reported in the original
publication. Dataset-scale statistics are retained as reported by the
corresponding papers and may therefore use different units (e.g., subjects,
recordings, or hours); they should not be interpreted as directly comparable
unless the measurement unit is explicitly matched.

\begin{table*}[!t]
\centering
\resizebox{\textwidth}{!}{
\begin{tabular}{lcccllll}
\toprule
Model &
Pretraining scale &
Data scale &
Parameters &
Input &
Pretraining objective &
Backbone &
Channel handling \\
\midrule
BENDR &
14,987 &
27,062 &
157.1M &
EEG &
Contrastive predictive coding &
Transformer &
Fixed 10--20 channels \\

BIOT &
NR &
NR &
3.4M &
EEG / ECG &
Contrastive learning &
Linear Transformer &
Fixed 18-channel bipolar montage \\

BrainOmni &
6,550 &
1,997 EEG + 656 MEG &
8.4M / 39.2M &
EEG / MEG &
Temporal masked reconstruction &
Criss-Cross Transformer &
Sensor encoder \\

Brant &
9 &
2,528 &
149.9M &
iEEG &
Masked autoencoding &
Transformer &
Subject-specific spatial encoder \\

CBraMod &
14,987 &
27,062 &
4.9M &
EEG &
Temporal masked reconstruction &
Criss-Cross Transformer &
Asymmetric conditional positional encoding \\

EEGMamba &
NR &
16,724 &
3.3M &
EEG &
Temporal masked reconstruction &
Mamba &
Spatiotemporal adaptive module \\

FEMBA &
14,987 &
27,062 &
7.8M / 46.6M / 77.8M / 389M &
EEG &
Temporal masked reconstruction &
Mamba &
Fixed 10--20 channels \\

LaBraM &
NR &
2,537.78 &
9.1M &
EEG &
Frequency-domain masked reconstruction &
Transformer &
Unified 10--20 channel vocabulary \\

NeuroGPT &
14,987 &
27,062 &
79.5M &
EEG &
Temporal masked reconstruction &
GPT-2 &
Fixed 10--20 channels \\

NeuroLM &
15,444 &
27,762 &
255.1M / 500M / 1.69B &
EEG &
Multi-task instruction tuning / autoregressive modeling &
Transformer &
Unified 10--20 channel vocabulary \\
\bottomrule
\end{tabular}
}
\caption{\textbf{Summary of pretrained neural-signal models considered in this study}.
NR denotes information not reported in the original paper. Pretraining-scale
statistics are reported using the terminology and units provided by the
corresponding publications.}
\label{tab:pretrained_model_comparison}
\end{table*}

\paragraph{BIOT \citep{yang2023biot}.}
BIOT (Biosignal Transformer) is a unified encoder for heterogeneous
biosignals, including EEG and ECG. It divides each channel into temporal
segments and converts channel--time patches into a shared token sequence.
A linear-attention Transformer then models dependencies among these tokens with
reduced computational cost. BIOT uses a contrastive learning objective and is
designed to facilitate transfer across datasets and signal modalities, although
its standard implementation assumes a fixed bipolar channel montage.

\paragraph{LaBraM\cite{jiang2024large}.}
LaBraM (Large Brain Model) is an EEG foundation model trained on heterogeneous
EEG data. It represents EEG recordings as sequences of channel-wise temporal
patches and incorporates channel identity through a unified vocabulary based on
the 10--20 system. Its pretraining objective reconstructs masked content in a
discretized neural-token space, with an emphasis on frequency-domain EEG
representations. LaBraM is designed to transfer across subjects, datasets, and
brain--computer interface paradigms.

\paragraph{NeuroLM\cite{jiang2025neurolm}.}
NeuroLM connects EEG representation learning with language-model-based task
generalization. EEG segments are encoded into neural tokens that can be
processed jointly with textual instructions. Through multitask instruction
tuning and autoregressive modeling, NeuroLM supports multiple downstream EEG
tasks within a shared model rather than requiring a separately designed output
head for each task. It uses a unified channel vocabulary to accommodate
heterogeneous EEG montages.

\paragraph{EEGPT\cite{gui2024eegmamba}.}
EEGPT is a pretrained Transformer for learning transferable EEG
representations from large-scale recordings. It jointly models within-channel
temporal dynamics and inter-channel spatial relationships, and is trained using
self-supervised reconstruction or prediction objectives over partially observed
EEG inputs. The pretrained encoder can subsequently be adapted to downstream
EEG classification and brain--computer interface tasks.

\paragraph{Brant\cite{zhang2023brant}.}
Brant is a foundation model for neural time-series representation learning,
with a primary focus on intracranial EEG recordings. It constructs tokens from
local signal segments and uses a Transformer encoder to capture long-range
temporal dependencies and cross-channel interactions. Its masked-autoencoding
pretraining objective is intended to learn representations that transfer across
recording sessions, subjects, and downstream tasks. In contrast to models
developed primarily for scalp EEG, Brant incorporates subject-specific spatial
encoding for high-resolution intracranial recordings.

Overall, the compared models represent several major directions in neural
foundation modeling. First, models such as BENDR and BIOT emphasize general
self-supervised representation learning and transfer across EEG or broader
biosignal datasets. Second, LaBraM, CBrAMod, EEGMamba, FEMBA, and NeuroGPT
explore masked reconstruction objectives with different architectural choices,
including Transformers, Criss-Cross attention, and state-space models.
Third, Brant extends large-scale neural representation learning to intracranial
recordings and incorporates subject-specific spatial information. Finally,
NeuroLM and BrainOmni investigate broader forms of generalization through,
respectively, language-conditioned multitask learning and multimodal neural
signal modeling. These differences in pretraining objective, backbone design,
and channel handling provide the basis for the representation-level comparison
in the main text.

\subsection{Unified Evaluation with Model-Specific Input Adapters}
\label{app:model_input_unification}

Although the compared pretrained models are evaluated under a unified
benchmark protocol, they are not assumed to receive an identical raw waveform
tensor. As documented in OmniEEG-Bench, different EEG foundation models impose
different input-format requirements, including target sampling rate, channel
handling, normalization, montage construction, and segment length. Some backbones are designed for flexible channel
configurations, whereas others assume fixed standard montages, bipolar
montages, zero-filled missing channels, or explicit channel identifiers.
Therefore, ``unified'' refers to the external benchmark interface and
downstream evaluation protocol rather than to a single model-agnostic input
tensor.

In practice, each EEG segment is first represented in a common channel-by-time
form,
\begin{equation}
    \mathbf{x}\in\mathbb{R}^{C\times T},
\end{equation}
after the shared preprocessing pipeline. A model-specific input adapter then
maps this representation to the format required by the selected backbone. This
adapter step may include resampling, channel selection or reordering, montage
conversion, temporal cropping or padding, and model-specific normalization.
After the backbone forward pass, token-, channel-, or window-level outputs are
converted into a fixed-dimensional feature matrix,
\begin{equation}
    \mathbf{F}\in\mathbb{R}^{B\times D_{\mathrm{flat}}},
\end{equation}
which is used by the same downstream probing or fine-tuning procedure. This
adapter-based design preserves each pretrained model's native input
assumptions while keeping downstream optimization, validation, model selection,
and test evaluation comparable across models.

\section{Details of Model Architecture}
\label{app:model_architecture}


This section provides a detailed description of the architecture used for EEG
pretraining. The model follows a Vision Transformer (ViT)-style design, in
which a multichannel EEG recording is represented as a sequence of
channel--time tokens. Each token corresponds to a temporal patch extracted
from a single EEG channel. The model consists of four main components:
(i) a patch tokenizer, (ii) learnable channel and temporal embeddings,
(iii) a Transformer encoder, and (iv) a token-wise time--frequency prediction
head. An overview of the default model configuration is provided in
Table~\ref{tab:model_configuration}.

\subsection{Channel--Time Token Representation}
\label{app:channel_time_tokens}

Let an EEG segment be divided into temporal patches independently for each
channel. Each resulting token represents one temporal patch from one EEG
channel. For a batch containing $B$ examples, the input to the model is
written as
\begin{equation}
    \mathbf{X}
    \in
    \mathbb{R}^{B \times S \times L},
\end{equation}
where $S$ is the number of channel--time tokens in each example and $L$
is the dimensionality of an individual input patch. When all channels and
temporal patches are retained, the sequence length is
\begin{equation}
    S = C T,
\end{equation}
where $C$ is the number of EEG channels and $T$ is the number of temporal
patches per channel. More generally, $S$ denotes the number of tokens
provided to the encoder after the token construction procedure.

In addition to the patch values, each token is associated with a channel
index and a temporal-patch index. These indices are represented as
\begin{equation}
    \mathbf{I}^{\mathrm{ch}}
    \in
    \{0,\ldots,C-1\}^{B \times S}
\end{equation}
and
\begin{equation}
    \mathbf{I}^{\mathrm{time}}
    \in
    \{0,\ldots,T-1\}^{B \times S},
\end{equation}
respectively. The two index tensors allow the model to distinguish patches
originating from different EEG channels and temporal locations.

\subsection{Patch Tokenizer}
\label{app:patch_tokenizer}

Each input patch is independently projected into the Transformer embedding
space using a shared patch tokenizer. Specifically, for the $s$-th token
of the $i$-th example, the initial token representation is computed as
\begin{equation}
    \widetilde{\mathbf{x}}_{i,s}
    =
    \operatorname{Dropout}
    \left(
        \operatorname{GELU}
        \left(
            \operatorname{LN}
            \left(
                \mathbf{W}_{\mathrm{tok}}
                \mathbf{x}_{i,s}
                +
                \mathbf{b}_{\mathrm{tok}}
            \right)
        \right)
    \right),
    \label{eq:patch_tokenizer}
\end{equation}
where
\begin{equation}
    \mathbf{W}_{\mathrm{tok}}
    \in
    \mathbb{R}^{D \times L},
    \qquad
    \mathbf{b}_{\mathrm{tok}}
    \in
    \mathbb{R}^{D},
\end{equation}
and $D$ denotes the Transformer embedding dimension. The tokenizer therefore
maps the input tensor from
$\mathbb{R}^{B \times S \times L}$ to
$\mathbb{R}^{B \times S \times D}$. The same tokenizer parameters are
shared across all channels and temporal positions.

\subsection{Channel and Temporal Embeddings}
\label{app:channel_temporal_embeddings}

To encode the spatial and temporal identity of each token, the model uses two
separate learnable embedding tables. The channel embedding table is
\begin{equation}
    \mathbf{E}^{\mathrm{ch}}
    \in
    \mathbb{R}^{C \times D},
\end{equation}
and the temporal embedding table is
\begin{equation}
    \mathbf{E}^{\mathrm{time}}
    \in
    \mathbb{R}^{T \times D}.
\end{equation}
For a token with channel index $c_{i,s}$ and temporal index $t_{i,s}$,
the input representation supplied to the Transformer encoder is
\begin{equation}
    \mathbf{z}^{(0)}_{i,s}
    =
    \widetilde{\mathbf{x}}_{i,s}
    +
    \mathbf{E}^{\mathrm{ch}}_{c_{i,s}}
    +
    \mathbf{E}^{\mathrm{time}}_{t_{i,s}}.
    \label{eq:token_embedding}
\end{equation}

The use of separate channel and temporal embeddings provides a factorized
representation of token position. In particular, the channel embedding
identifies the spatial origin of a token, whereas the temporal embedding
identifies its location within the temporal sequence. This formulation also
avoids assigning an independent positional embedding to every possible
channel--time combination.

No additional classification token is appended to the sequence. Instead, all
channel--time tokens are passed directly to the Transformer, and the model
produces a prediction for every token.

\subsection{Transformer Encoder}
\label{app:transformer_encoder}

The position-aware token sequence
\begin{equation}
    \mathbf{Z}^{(0)}
    \in
    \mathbb{R}^{B \times S \times D}
\end{equation}
is processed by a stack of $N$ Transformer encoder layers. Each layer uses
multi-head self-attention followed by a position-wise feed-forward network.
The encoder adopts a pre-normalization design, meaning that layer
normalization is applied before both the self-attention and feed-forward
submodules.

For the $\ell$-th Transformer layer, the self-attention sublayer can be
expressed as
\begin{equation}
    \mathbf{U}^{(\ell)}
    =
    \mathbf{Z}^{(\ell-1)}
    +
    \operatorname{Dropout}
    \left(
        \operatorname{MSA}
        \left(
            \operatorname{LN}
            \left(
                \mathbf{Z}^{(\ell-1)}
            \right)
        \right)
    \right),
    \label{eq:transformer_attention}
\end{equation}
where $\operatorname{MSA}(\cdot)$ denotes multi-head self-attention. The
feed-forward sublayer is then given by
\begin{equation}
    \mathbf{Z}^{(\ell)}
    =
    \mathbf{U}^{(\ell)}
    +
    \operatorname{Dropout}
    \left(
        \operatorname{FFN}
        \left(
            \operatorname{LN}
            \left(
                \mathbf{U}^{(\ell)}
            \right)
        \right)
    \right).
    \label{eq:transformer_ffn}
\end{equation}

For $H$ attention heads, the output of multi-head self-attention is
\begin{equation}
    \operatorname{MSA}(\mathbf{Z})
    =
    \operatorname{Concat}
    \left(
        \operatorname{head}_{1},
        \ldots,
        \operatorname{head}_{H}
    \right)
    \mathbf{W}^{O},
\end{equation}
where each attention head is calculated as
\begin{equation}
    \operatorname{head}_{h}
    =
    \operatorname{Softmax}
    \left(
        \frac{
            \mathbf{Q}_{h}\mathbf{K}_{h}^{\top}
        }{
            \sqrt{D_{h}}
        }
    \right)
    \mathbf{V}_{h}.
    \label{eq:self_attention}
\end{equation}
Here,
\begin{equation}
    \mathbf{Q}_{h}=\mathbf{Z}\mathbf{W}^{Q}_{h},
    \qquad
    \mathbf{K}_{h}=\mathbf{Z}\mathbf{W}^{K}_{h},
    \qquad
    \mathbf{V}_{h}=\mathbf{Z}\mathbf{W}^{V}_{h},
\end{equation}
and $D_h=D/H$ is the dimensionality of each attention head. Self-attention
is performed across the complete channel--time token sequence, allowing the
model to capture both cross-channel dependencies and long-range temporal
interactions.

The position-wise feed-forward network consists of two linear transformations
with a GELU activation:
\begin{equation}
    \operatorname{FFN}(\mathbf{z})
    =
    \mathbf{W}_{2}
    \operatorname{GELU}
    \left(
        \mathbf{W}_{1}\mathbf{z}
        +
        \mathbf{b}_{1}
    \right)
    +
    \mathbf{b}_{2},
    \label{eq:ffn}
\end{equation}
where the hidden dimensionality of the feed-forward network is
\begin{equation}
    D_{\mathrm{ff}}
    =
    rD,
\end{equation}
and $r$ is the MLP expansion ratio.

Under the default configuration, the Transformer has $N=12$ encoder
layers, an embedding dimension of $D=512$, and $H=8$ attention heads.
Consequently, each attention head has a dimensionality of
\begin{equation}
    D_h = \frac{512}{8} = 64.
\end{equation}
The MLP expansion ratio is $r=4$, resulting in a feed-forward hidden
dimension of
\begin{equation}
    D_{\mathrm{ff}} = 4 \times 512 = 2048.
\end{equation}

\subsection{Token-Wise Time--Frequency Prediction Head}
\label{app:time_frequency_head}

The output of the final Transformer layer is
\begin{equation}
    \mathbf{Z}^{(N)}
    \in
    \mathbb{R}^{B \times S \times D}.
\end{equation}
A shared prediction head is applied independently to every channel--time
token. The head first transforms each token representation within the model
embedding space and then projects it to the desired time--frequency output
dimension:
\begin{equation}
    \mathbf{o}_{i,s}
    =
    \mathbf{W}_{\mathrm{out}}
    \operatorname{GELU}
    \left(
        \mathbf{W}_{\mathrm{h}}
        \mathbf{z}^{(N)}_{i,s}
        +
        \mathbf{b}_{\mathrm{h}}
    \right)
    +
    \mathbf{b}_{\mathrm{out}},
    \label{eq:prediction_head}
\end{equation}
where
\begin{equation}
    \mathbf{W}_{\mathrm{h}}
    \in
    \mathbb{R}^{D \times D}
\end{equation}
is the hidden projection and
\begin{equation}
    \mathbf{W}_{\mathrm{out}}
    \in
    \mathbb{R}^{(FK) \times D}
\end{equation}
is the output projection. Here, $F$ denotes the number of frequency bands
and $K$ denotes the number of temporal frames associated with each input
patch.

The direct output of the prediction head has the shape
\begin{equation}
    \mathbf{O}
    \in
    \mathbb{R}^{B \times S \times (FK)}.
\end{equation}
It is subsequently reshaped into
\begin{equation}
    \widehat{\mathbf{Y}}
    =
    \operatorname{Reshape}(\mathbf{O})
    \in
    \mathbb{R}^{B \times S \times F \times K}.
    \label{eq:prediction_reshape}
\end{equation}
Thus, for every input channel--time token, the model predicts a
frequency-band energy trajectory containing $F$ frequency bands and $K$
temporal frames. The prediction-head parameters are shared across all tokens.

\subsection{Default Architecture Configuration}
\label{app:default_architecture}

Table~\ref{tab:model_configuration} summarizes the default hyperparameters
used to construct the model. Parameters associated with input construction
and prediction targets, including $L$, $C$, $T$, $F$, and $K$, are
determined by the preprocessing and target-generation procedures.

\begin{table}[t]
    \centering
    \small
    \setlength{\tabcolsep}{5pt}
    \renewcommand{\arraystretch}{1.12}
    \begin{tabular}{lc}
        \toprule
        \textbf{Component / hyperparameter}
        & \textbf{Value} \\
        \midrule
        Token embedding dimension $D$
        & 512 \\
        Number of Transformer layers $N$
        & 12 \\
        Number of attention heads $H$
        & 8 \\
        Dimension per attention head $D_h$
        & 64 \\
        MLP expansion ratio $r$
        & 4.0 \\
        Feed-forward dimension $D_{\mathrm{ff}}$
        & 2,048 \\
        Dropout probability
        & 0.1 \\
        Transformer normalization
        & Pre-LN \\
        Transformer activation
        & GELU \\
        Positional representation
        & Channel + temporal embeddings \\
        Output strategy
        & Token-wise prediction \\
        \bottomrule
    \end{tabular}
    \caption{Default configuration of the EEG pretraining model.}
    \label{tab:model_configuration}
\end{table}

\subsection{1B-Parameter Model Configuration}
\label{app:model_configuration_1B}

For the large-scale pretraining experiment, we construct a 1B-parameter
variant of the EEG pretraining model. The large-scale model retains the same
overall architecture as the default configuration, including the shared patch
tokenizer, factorized channel and temporal embeddings, pre-normalized
Transformer encoder, and token-wise time--frequency prediction head. The
model capacity is increased primarily by scaling the width and depth of the
Transformer encoder.

Specifically, the embedding dimension is increased from $D=512$ to
$D=2048$, and the number of Transformer encoder layers is increased from
$N=12$ to $N=24$. The number of attention heads is increased from $H=8$ to
$H=16$. Consequently, the dimensionality of each attention head is
\begin{equation}
D_h = \frac{D}{H}
= \frac{2048}{16}
= 128.
\end{equation}
The MLP expansion ratio remains fixed at $r=4$, resulting in a feed-forward
hidden dimension of
\begin{equation}
D_{\mathrm{ff}}
= rD
= 4 \times 2048
= 8192.
\end{equation}

All other architectural choices, including the use of GELU activations,
pre-layer normalization, a dropout probability of $0.1$, and separate
learnable channel and temporal embeddings, remain unchanged from the default
model. The complete configuration of the 1B-parameter model is summarized in
Table~\ref{tab:model_configuration_1B}.

\begin{table}[t]
    \centering
    \small
    \setlength{\tabcolsep}{5pt}
    \renewcommand{\arraystretch}{1.12}
    \begin{tabular}{lc}
        \toprule
        \textbf{Component / hyperparameter}
        & \textbf{Value} \\
        \midrule
        Token embedding dimension $D$
        & 2048 \\
        Number of Transformer layers $N$
        & 24 \\
        Number of attention heads $H$
        & 16 \\
        Dimension per attention head $D_h$
        & 128 \\
        MLP expansion ratio $r$
        & 4.0 \\
        Feed-forward dimension $D_{\mathrm{ff}}$
        & 8192 \\
        Dropout probability
        & 0.1 \\
        Transformer normalization
        & Pre-LN \\
        Transformer activation
        & GELU \\
        Positional representation
        & Channel + temporal embeddings \\
        Output strategy
        & Token-wise prediction \\
        \bottomrule
    \end{tabular}
    \caption{Configuration of the 1B-parameter EEG pretraining model.}
    \label{tab:model_configuration_1B}
\end{table}

\subsection{Overall Forward Process}
\label{app:model_forward_process}

The complete forward process can be summarized as
\begin{align}
    \widetilde{\mathbf{X}}
    &=
    \operatorname{Tokenizer}(\mathbf{X}),
    \\
    \mathbf{Z}^{(0)}
    &=
    \widetilde{\mathbf{X}}
    +
    \operatorname{Embed}_{\mathrm{ch}}
    \left(
        \mathbf{I}^{\mathrm{ch}}
    \right)
    +
    \operatorname{Embed}_{\mathrm{time}}
    \left(
        \mathbf{I}^{\mathrm{time}}
    \right),
    \\
    \mathbf{Z}^{(N)}
    &=
    \operatorname{TransformerEncoder}
    \left(
        \mathbf{Z}^{(0)}
    \right),
    \\
    \widehat{\mathbf{Y}}
    &=
    \operatorname{Reshape}
    \left(
        \operatorname{PredictionHead}
        \left(
            \mathbf{Z}^{(N)}
        \right)
    \right).
\end{align}

Accordingly, the model implements the following shape transformation:
\begin{equation}
    \underbrace{
        \mathbb{R}^{B \times S \times L}
    }_{\text{input patches}}
    \longrightarrow
    \underbrace{
        \mathbb{R}^{B \times S \times D}
    }_{\text{token representations}}
    \longrightarrow
    \underbrace{
        \mathbb{R}^{B \times S \times F \times K}
    }_{\text{time--frequency predictions}}.
\end{equation}

This architecture preserves the identity of every channel--time patch
throughout the network. At the same time, global self-attention allows each
token representation to incorporate information from all other channels and
temporal locations before the token-wise prediction is produced.

\section{Selection of checkpoints and representation layers}


To establish a consistent representation extraction protocol, we first investigated how the choice of Transformer layer affects downstream performance. We used the final pretraining checkpoint (after five epochs) and extracted embeddings from each Transformer block. Each layer's representation was evaluated using the linear-probing protocol on the ADHD classification task. The results showed that representations from the ninth Transformer block, corresponding to approximately 80\% of the network depth, achieved the best performance. Therefore, we selected the 80\%-depth layer as the default representation layer for all subsequent experiments.

After fixing the representation layer, we further examined the effect of pretraining checkpoints. We evaluated embeddings extracted from the selected layer across checkpoints saved after each training epoch using the same linear-probing protocol. Linear-probing performance consistently improved throughout pretraining, and the checkpoint after the fifth epoch achieved the best performance. Consequently, we used the fifth-epoch checkpoint as the final pretrained model for all downstream evaluations.

This layer and checkpoint selection procedure was performed before downstream experiments and was applied consistently to all model variants, including both the 50M and 1B parameter models.

\subsection{Fine-Tuning Protocol for Downstream Tasks}
\label{app:downstream_finetuning}

After pretraining, we adapt the pretrained model to each downstream classification task using a unified fine-tuning protocol. Given an input sample, we extract the token representations from the ninth layer of the pretrained encoder. Let $B$, $N$, and $D$ denote the batch size, the number of tokens, and the token embedding dimension, respectively. The extracted representations are therefore given by

\begin{equation}
    \mathbf{H}^{(9)} \in \mathbb{R}^{B \times N \times D}.
\end{equation}

To obtain a fixed-dimensional representation for downstream classification, we flatten the token and embedding dimensions:

\begin{equation}
    \mathbf{h}
    =
    \operatorname{Flatten}\left(\mathbf{H}^{(9)}\right)
    \in
    \mathbb{R}^{B \times (ND)}.
\end{equation}

This operation concatenates the representations of all tokens while preserving their original order, rather than aggregating them through mean pooling or selecting only a designated token.

A newly initialized linear classification head is then applied to the flattened representation. For a downstream task with $K$ classes, the prediction logits are computed as

\begin{equation}
    \mathbf{Z}
    =
    \mathbf{h}\mathbf{W}^{\top} + \mathbf{b},
\end{equation}

where

\begin{equation}
    \mathbf{W} \in \mathbb{R}^{K \times ND},
    \qquad
    \mathbf{b} \in \mathbb{R}^{K},
\end{equation}

are the learnable weight matrix and bias vector of the task-specific classification head, respectively. The predicted class probabilities are obtained using the softmax function:

\begin{equation}
    p(y=k \mid \mathbf{h})
    =
    \frac{\exp(Z_k)}
    {\sum_{j=1}^{K}\exp(Z_j)}.
\end{equation}

The model is trained by minimizing the cross-entropy loss:

\begin{equation}
    \mathcal{L}_{\mathrm{CE}}
    =
    -\frac{1}{B}
    \sum_{i=1}^{B}
    \sum_{k=1}^{K}
    y_{i,k}\log p(y_i=k \mid \mathbf{h}_i),
\end{equation}
where $y_{i,k}$ is the one-hot encoded ground-truth label for the $i$-th sample.

For each downstream dataset, we divide the available samples into training, validation, and test sets using an $8{:}1{:}1$ ratio. The training set is used for parameter optimization, whereas the validation set is used for model selection. During training, the model is periodically evaluated on the validation set, and the checkpoint achieving the best validation performance is retained. After model selection, the selected checkpoint is evaluated on the held-out test set, and the resulting performance is reported as the final downstream result. The test set is not used for parameter optimization, hyperparameter tuning, or checkpoint selection. Unless otherwise specified, the same data-splitting, fine-tuning, and model-selection protocol is consistently applied to all downstream tasks.

\section{Performance Comparison between FAME and MAE}

To further evaluate the effectiveness of frequency-balanced reconstruction, we
conduct a controlled comparison between FAME and the conventional MAE
objective. Specifically, we pretrain a baseline model using the standard MAE
objective on the same pretraining dataset, with the same model size (50M
parameters), training configuration, and random seed as FAME. The pretrained
models are then evaluated on all downstream tasks using the same linear probing
protocol.

The only architectural difference between the two models lies in the final
reconstruction head. Since MAE directly reconstructs the raw EEG signal,
whereas FAME reconstructs frequency-normalized targets with balanced
frequency-band supervision, their output dimensions and corresponding
reconstruction heads are different. Apart from this task-specific output layer,
all other architectural components and optimization hyperparameters are kept
identical to ensure a fair comparison.

\begin{figure*}[!t]
  \centering
  \includegraphics[width=0.99\linewidth]{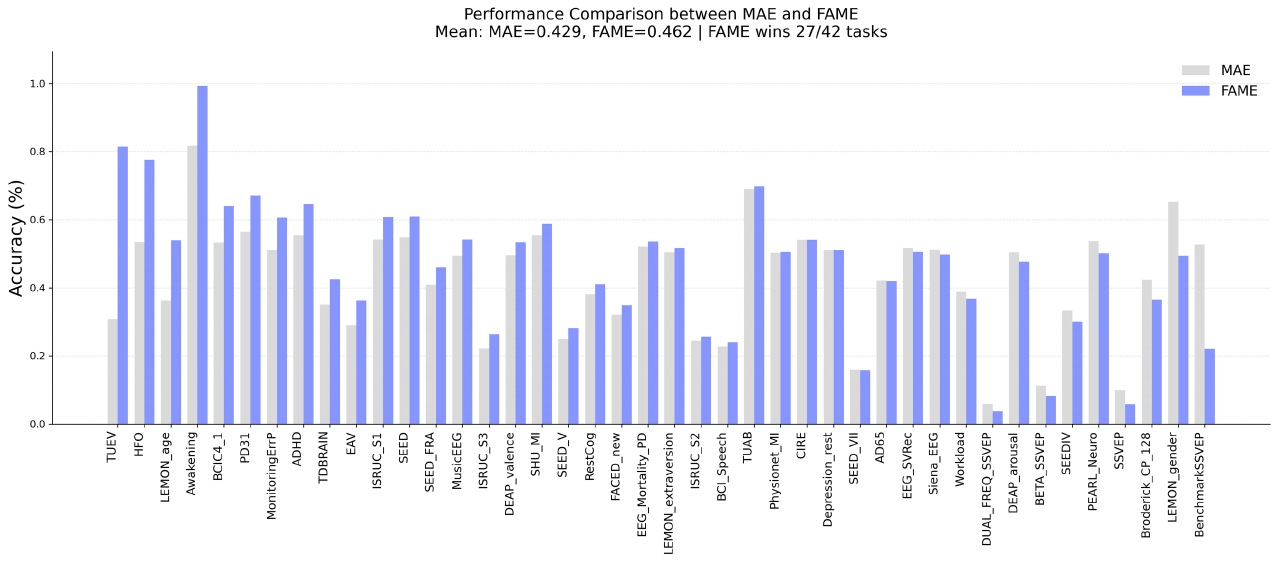}
  \caption{
    \textbf{Performance comparison between FAME and MAE across downstream EEG
    tasks.}
    The tasks are sorted according to the performance difference between FAME and
    MAE, with tasks showing larger improvements placed on
    the left. FAME consistently outperforms MAE on a broad range of downstream
    benchmarks, achieving better performance on 27 out of 42 tasks.
    }
  \label{fig:MAEPerformanceComparison}
\end{figure*}

As shown in Figure~\ref{fig:MAEPerformanceComparison}, FAME consistently
improves downstream performance compared with standard MAE. Across 42
downstream tasks, FAME achieves better performance on 27 tasks, while achieving
comparable results on most of the remaining tasks. Notably, FAME provides
substantial improvements on several challenging EEG understanding tasks,
including TUEV and HFO, suggesting that frequency-balanced reconstruction
enables the model to learn more transferable representations beyond the
dominant low-frequency components captured by conventional MAE objectives.
Overall, these results demonstrate that the proposed frequency-balanced
supervision provides a more effective pretraining objective for EEG
representation learning.

\section{Details of Linear Probe of 1B Model}
\label{app:linearProbe_detail}

\subsection{Multitask Linear Probing}
\label{sec:multitask_linear_probe}

To evaluate the quality of representations learned during pretraining, we perform multitask linear probing on a diverse collection of downstream EEG classification datasets. Instead of updating the pretrained encoder, we keep the entire EEG Transformer backbone frozen and train lightweight task-specific prediction heads on top of extracted representations.

The evaluation includes five groups of downstream tasks, covering disease-related diagnosis, consciousness state recognition, cognition and emotion analysis, natural stimulus decoding, and motor interaction. The initial configuration contains 48 datasets. Two datasets (\textit{HBN\_EEG} and \textit{ChineseEEG2\_RA\_Tone}) are excluded because their processed inputs do not contain valid EEG tokens under the current tokenization scheme, resulting in 46 effective downstream tasks.

\subsection{Representation Extraction}

The pretrained EEG foundation model contains 20 Transformer blocks with a hidden dimension of 2048, 16 attention heads, and an MLP expansion ratio of 4.0. For linear probing, we extract intermediate representations rather than using the final Transformer output. Specifically, we select the output of the 15th Transformer block (layer index 14), corresponding to 75\% of the model depth.

The pretrained encoder is used as a frozen feature extractor, and no gradient is propagated through the patch tokenizer, channel embeddings, temporal embeddings, or Transformer layers. Given an EEG segment, the encoder produces token-level representations. Invalid tokens corresponding to unavailable EEG channels are removed according to the token validity mask, and the remaining token embeddings are concatenated in their original order.

For a dataset with $C$ valid EEG channels and 10 temporal patches per channel, the resulting feature dimension is $C \times 10 \times 2048$. These extracted representations are then used as inputs to independent task-specific classifiers.

\subsection{Task-specific Classification Heads}

Each downstream dataset is equipped with an independent classification head while sharing the same frozen pretrained representation. The classification head consists of LayerNorm, dropout, and a linear classifier:

\[
h(x)=W \operatorname{Dropout}(\operatorname{LayerNorm}(x))+b .
\]

The classifier parameters are optimized independently across all downstream tasks, while the pretrained encoder remains unchanged. This design evaluates the intrinsic transferability of the learned EEG representations without introducing additional adaptation of the backbone.

\subsection{Optimization Procedure}

All linear probing experiments are trained for 10 epochs using distributed data parallel training on four GPUs. The effective batch size is 64 (16 samples per GPU) without gradient accumulation. We optimize only the task-specific classification heads using AdamW with a learning rate of $10^{-3}$ and weight decay of $10^{-3}$.

The learning rate is scheduled using cosine annealing with a minimum learning rate of $10^{-5}$. No class re-weighting, balanced sampling, focal loss, warm-up, or gradient accumulation is applied. The training objective is the standard cross-entropy loss.

Since different downstream datasets contain different numbers of training samples, multitask optimization follows a round-robin scheduling strategy. Each batch from one task produces an independent optimization step for its corresponding classification head. The pretrained encoder is shared across all tasks but remains frozen throughout training.

\subsection{Subject-level Data Splitting}

To avoid subject leakage, all datasets are split at the subject level whenever subject identifiers are available. Each dataset is divided into training, validation, and test sets using an 80/10/10 split.

For datasets where subjects contain samples from multiple classes, we perform group-aware multilabel splitting, assigning all samples from the same subject to the same split while approximately preserving class distributions. For datasets where each subject belongs to a single class, stratified subject-level splitting is applied to maintain class coverage across subsets.

When reliable subject-level splitting cannot be performed because of missing metadata or insufficient subjects, we use a predefined fallback split and record the corresponding splitting strategy.

\subsection{Evaluation and Model Selection}

For each downstream task, the best classifier checkpoint is selected according to validation balanced accuracy. The frozen encoder is shared across all tasks, while each task maintains its own optimal classification head.

We report accuracy, balanced accuracy, macro-averaged metrics, per-class accuracy, and confusion matrices. Because the backbone representation remains fixed, differences across downstream tasks primarily reflect the quality and transferability of the pretrained EEG representations.

\section{Additional Evaluation on NeuralBench}

To further evaluate the generalization capability of FAME beyond our primary
evaluation suite, we benchmark the pretrained models on NeuralBench, a large
collection of heterogeneous EEG downstream tasks spanning cognitive,
clinical, and brain-computer interface applications. Different from our
linear-probing evaluation, NeuralBench adopts a full fine-tuning protocol,
where all model parameters are updated during downstream adaptation. We
compare FAME against several representative EEG foundation models, including
BENDR, BIOT, CBraMod, EEGNet, LaBraM, LUNA, and REVE, following the official
NeuralBench evaluation protocol.

As shown in Table~\ref{tab:downstream_linear_probe_neuralBench}, FAME-50M achieves the
highest average performance across the evaluated NeuralBench tasks, obtaining
a mean balanced accuracy of 61.4\%. With only 50M parameters, FAME-50M
outperforms larger EEG foundation models and achieves the best performance on
several individual benchmarks, including mental arithmetic decoding and N2PC
classification. It also remains highly competitive on clinical tasks such as
depression diagnosis. These results demonstrate that the benefits of
frequency-balanced pretraining extend beyond linear probing: the learned
representations remain effective when the entire model is adapted to diverse
downstream objectives, highlighting the robustness and transferability of
FAME representations.

We further evaluate a scaled-up FAME model with approximately 1B parameters.
Although the larger model achieves improvements on several individual tasks,
including audiovisual stimulus classification, depression diagnosis, and
emotion recognition, its overall average performance is slightly lower than
that of FAME-50M. This observation suggests that increasing model capacity
does not necessarily translate into uniform improvements under full
fine-tuning settings. One possible explanation is that adapting larger EEG
foundation models requires more careful optimization and task-specific
regularization, as the increased parameter space introduces additional
challenges during downstream fine-tuning. Overall, these NeuralBench results
provide further evidence that FAME is a robust EEG foundation model, and that
frequency-balanced pretraining enables transferable representations across
heterogeneous EEG applications.

\definecolor{best}{RGB}{255,220,238}
\definecolor{famelarge}{RGB}{255,220,238}

\begin{table*}[t]
\centering
\label{tab:downstream_comparison}
\resizebox{\textwidth}{!}{
\begin{tabular}{l|ccccccc|cc}
\toprule
\textbf{Task} & \textbf{EEGNet} & \textbf{CBraMod} & \textbf{BIOT} & \textbf{LaBraM} & \textbf{LUNA} & \textbf{BENDR} & \textbf{REVE} & \textbf{FAME-50M} & \textbf{FAME-1B} \\
\midrule
Audiovisual Stimulus 
& 23.6$\pm$1.8 & 34.6$\pm$9.1 & 25.8$\pm$3.2 & 39.9$\pm$2.8 
& 27.0$\pm$3.9 & 44.9$\pm$0.9 & 48.9$\pm$5.4 & 44.5$\pm$5.1 & \cellcolor{best}\textbf{57.7$\pm$4.1} \\

Mental Arithmetic 
& 50.3$\pm$2.7 & 70.0$\pm$4.9 & 64.0$\pm$3.4 & 68.3$\pm$5.2 
& 69.6$\pm$1.4 & 62.2$\pm$5.8 & 71.6$\pm$0.9 & \cellcolor{best}\textbf{74.3$\pm$5.9} & 72.9$\pm$4.4 \\

Depression Diagnosis 
& 81.9$\pm$3.6 & 85.9$\pm$1.7 & 81.2$\pm$6.9 & 86.3$\pm$3.5 
& 84.7$\pm$1.7 & 86.9$\pm$0.7 & 86.0$\pm$0.8 & 86.1$\pm$4.1 & \cellcolor{best}\textbf{88.4$\pm$3.3} \\

Dementia Diagnosis 
& 44.2$\pm$4.1 & 38.9$\pm$5.8 & \cellcolor{best}\textbf{45.3$\pm$0.3} & 41.5$\pm$0.7 
& 42.1$\pm$0.9 & 29.6$\pm$4.9 & 42.1$\pm$4.6 & 45.2$\pm$0.8 & 39.6$\pm$2.7 \\

Lrp 
& 74.0$\pm$0.4 & 78.0$\pm$0.2 & 50.1$\pm$0.3 & 77.9$\pm$0.5 
& 74.5$\pm$0.6 & 70.5$\pm$3.2 & \cellcolor{best}\textbf{80.8$\pm$1.3} & 76.8$\pm$0.4 & 71.1$\pm$0.5 \\

Mismatch Negativity 
& 58.6$\pm$0.8 & \cellcolor{best}\textbf{58.9$\pm$0.3} & 50.0$\pm$0.1 & 58.6$\pm$0.3 
& 57.7$\pm$0.7 & 50.2$\pm$0.1 & 57.7$\pm$1.8 & 58.6$\pm$0.8 & 52.2$\pm$2.4 \\

N170 
& 68.9$\pm$0.9 & 71.3$\pm$1.1 & 50.1$\pm$0.2 & 73.0$\pm$0.4 
& 68.8$\pm$0.7 & 67.2$\pm$1.1 & \cellcolor{best}\textbf{76.5$\pm$1.9} & 72.4$\pm$0.9 & 68.5$\pm$1.6 \\

N2pc 
& 62.0$\pm$0.6 & 62.0$\pm$0.8 & 50.0$\pm$0.1 & 61.6$\pm$1.2 
& 62.1$\pm$1.2 & 56.3$\pm$0.1 & 62.4$\pm$0.8 & \cellcolor{best}\textbf{64.9$\pm$1.2} & 60.2$\pm$3.0 \\

N400 
& \cellcolor{best}\textbf{64.6$\pm$1.5} & 61.1$\pm$0.5 & 49.7$\pm$0.5 & 64.1$\pm$1.6 
& 63.2$\pm$0.2 & 60.8$\pm$2.5 & 63.8$\pm$0.7 & 61.7$\pm$0.7 & 61.2$\pm$1.2 \\

SSVEP 
& 8.5$\pm$0.9 & 80.1$\pm$9.6 & 58.4$\pm$1.9 & \cellcolor{best}\textbf{96.6$\pm$0.7} 
& 12.6$\pm$0.4 & 9.7$\pm$4.2 & \cellcolor{best}\textbf{96.6$\pm$0.8} & 95.9$\pm$1.2 & 93.2$\pm$3.0 \\

Mental Imagery 
& \cellcolor{best}\textbf{29.6$\pm$1.7} & 20.6$\pm$1.1 & 26.6$\pm$1.0 & 21.2$\pm$0.3 
& 21.2$\pm$0.3 & 21.8$\pm$0.7 & 26.3$\pm$0.5 & 21.4$\pm$1.0 & 23.3$\pm$0.5 \\

Schizophrenia Diagnosis 
& \cellcolor{best}\textbf{71.0$\pm$4.9} & 54.0$\pm$2.8 & 55.9$\pm$3.6 & 61.9$\pm$1.8 
& 64.5$\pm$0.6 & 57.9$\pm$0.2 & 47.2$\pm$5.0 & 67.7$\pm$3.0 & 64.7$\pm$4.0 \\

Emotion 
& 23.2$\pm$0.9 & 30.5$\pm$2.6 & 12.2$\pm$1.3 & 26.7$\pm$0.5 
& 32.5$\pm$10.1 & 12.7$\pm$1.6 & 31.7$\pm$2.1 & 28.6$\pm$0.1 & \cellcolor{best}\textbf{37.3$\pm$0.7} \\

\midrule
Average score 
& 50.8 & 57.4 & 47.6 & 59.8 & 52.3 & 48.5 & 60.9 & \cellcolor{best}\textbf{61.4} & 60.8 \\
\bottomrule
\end{tabular}
}
\caption{Performance comparison on EEG downstream tasks. Results are reported as balanced accuracy (\%). The best performance for each task is highlighted in bold.}
\label{tab:downstream_linear_probe_neuralBench}
\end{table*}

\section{Computational Resources}
\label{sec:computational_resources}

Pretraining was performed on eight NVIDIA H20 GPUs, each equipped with
96~GB of HBM3 memory (768~GB total). The training system provided
140~vCPUs and 1960~GiB of system memory. We trained the model for five
epochs using all eight GPUs. A complete pretraining run required
approximately 12 hours of wall-clock time, corresponding to approximately 96 GPU-hours. The reported runtime covers pretraining only.

\begin{figure*}[!t]
  \centering
  \includegraphics[width=0.99\linewidth]{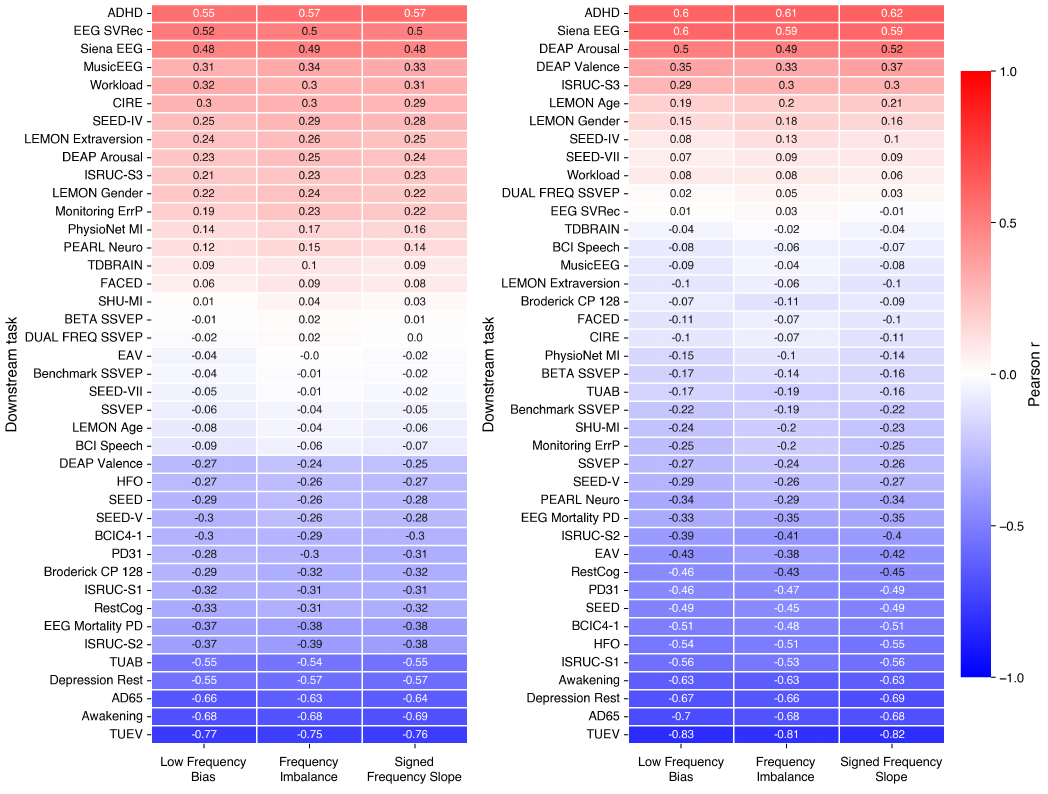}
  \caption{
  \textbf{Association between frequency bias and downstream generalization across EEG tasks.}
  Each heatmap shows the correlation between three model-level frequency-bias metrics (columns) and linear-probing performance across downstream tasks (rows). Results are computed without FAME (left) and with FAME (right). The overall patterns are similar in both settings, suggesting that the observed associations are not driven solely by FAME and vary across tasks.
  }
  \label{fig:appendix_frequencyBias_generalization}
\end{figure*}

\begin{figure*}[!t]
  \centering
  \includegraphics[width=0.99\linewidth]{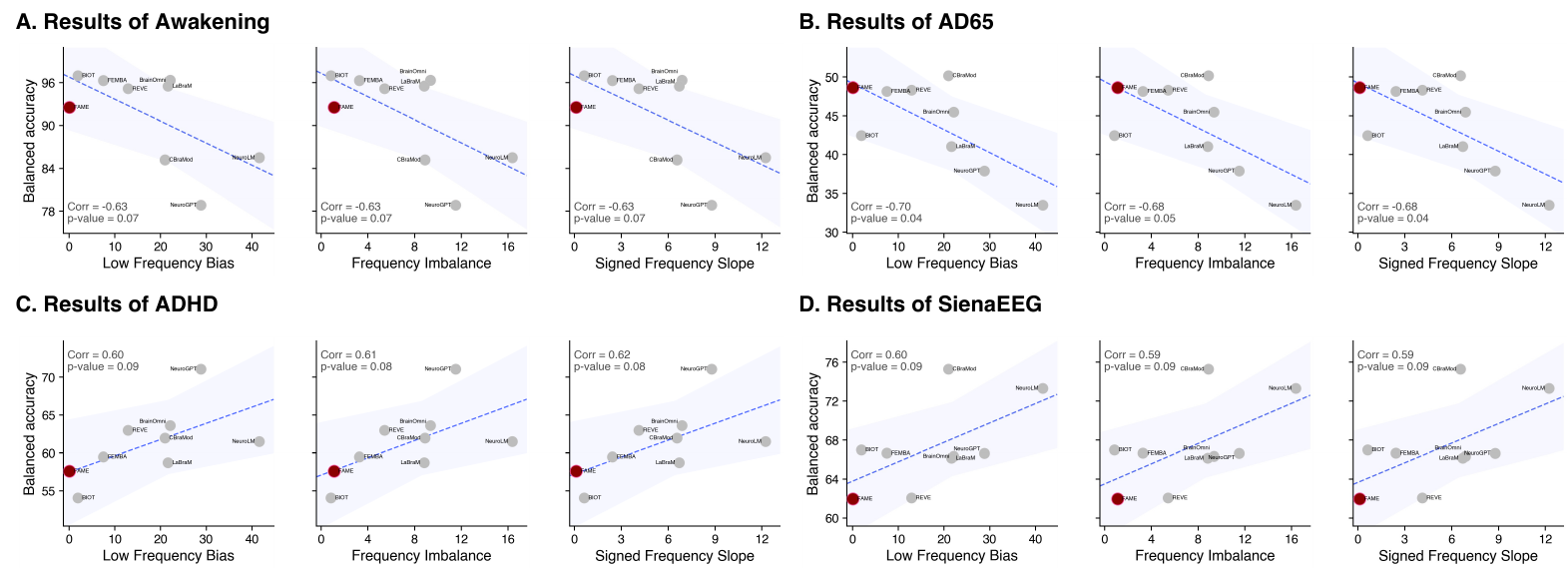}
  \caption{
  \textbf{Relationship between frequency bias and downstream generalization performance.}
  We visualize the association between the frequency bias metric and downstream
  performance across four representative EEG tasks.
  }
  \label{fig:appendix_frequencyBias_scatter}
\end{figure*}

\section{Additional Results on the Association Between Frequency Bias and Downstream Generalization}
\label{sec:appendix_frequency_bias_generalization}

Figure~\ref{fig:appendix_frequencyBias_generalization} presents the complete correlation results between representational frequency bias and downstream linear-probing performance. Each row corresponds to a downstream EEG task, and the three columns report results obtained using three complementary frequency-bias metrics: the high-to-low frequency reconstruction-loss ratio, the frequency-imbalance metric, and the spectrum slope. To assess whether the observed associations are sensitive to the inclusion of our frequency-balanced model, we report correlations both without FAME (left) and with FAME (right). In addition to these aggregated correlation analyses, Figure~\ref{fig:appendix_frequencyBias_scatter} provides task-level scatter plots for four representative downstream tasks, illustrating the detailed relationship between frequency bias and generalization performance.

Across both settings, frequency bias is associated with downstream performance on multiple tasks. In particular, consistent relationships are observed for TUEV disease classification and AD65 classification. Models that preserve information more evenly across frequency bands generally achieve stronger linear-probing performance on these tasks. This observation suggests that successful transfer to clinical classification benefits from representations that retain complementary information distributed across the EEG spectrum, rather than predominantly preserving activity within a restricted frequency range.

The direction and magnitude of the association nevertheless vary across downstream tasks. For ADHD classification, for example, better relative preservation of high-frequency information is associated with stronger downstream performance. This task dependence is plausible because discriminative EEG signatures are not uniformly distributed across frequencies: whereas some tasks may require information integrated across a broad spectral range, others may rely more strongly on activity within particular frequency bands. Consequently, frequency balance should not be interpreted as requiring identical information preservation at every frequency, but as reducing a systematic representational preference that could discard task-relevant spectral information.

Overall, these results show that the relationship between frequency-dependent information preservation and downstream generalization is widespread but task dependent. In many tasks, reduced frequency bias is associated with improved transfer, while a smaller subset benefits from preferential preservation of particular spectral components. These findings support frequency balance as a useful model-level indicator of representation quality, while also highlighting the importance of considering the spectral characteristics of each downstream task.

\end{document}